\documentclass{article}

\usepackage{arxiv}

\usepackage[utf8]{inputenc} 
\usepackage[T1]{fontenc}    
\usepackage{hyperref}       
\usepackage{url}            
\usepackage{booktabs}       
\usepackage{amsfonts}       
\usepackage{nicefrac}       
\usepackage{microtype}      
\usepackage{lipsum}		
\usepackage{graphicx}
\usepackage{natbib}
\usepackage{doi}
\usepackage{graphicx}%
\usepackage{multirow}%
\usepackage{amsmath,amssymb,amsfonts}%
\usepackage{amsthm}%
\usepackage{mathrsfs}%
\usepackage[title]{appendix}%
\usepackage{xcolor}%
\usepackage{textcomp}%
\usepackage{manyfoot}%
\usepackage{booktabs}%
\usepackage{algorithm}%
\usepackage{algorithmicx}%
\usepackage{algpseudocode}%
\usepackage{listings}%
\usepackage{makecell}
\usepackage{amssymb}
\usepackage{natbib}
\usepackage{xcolor}
\usepackage{tikz}
\usetikzlibrary{calc, intersections, arrows, shapes, positioning, shadows, trees}
\usepackage{amsmath}
\newcommand{\varB}[1]{{\operatorname{\mathit{#1}}}}
\usepackage{listings}

\usepackage{pifont}

\newcommand{\tikzcircle}[2][red,fill=red]{\tikz[baseline=-0.5ex]\draw[#1,radius=#2] (0,0) circle ;}%
\newcommand{\tikzsquare}[2][red,fill=red]{\tikz[baseline=-0.5ex]\draw[#1,radius=#2] (-0.1, -0.1) rectangle (0.1,0.1);}%
\newcommand{\tikztriangle}[2][red,fill=red]{\tikz[baseline=-0.5ex]\draw[#1,radius=#2] (-0.1,-0.1) -- (0,0.1) -- (0.1,-0.1);}%
\newcommand{\tikzinvtriangle}[2][red,fill=red]{\tikz[baseline=-0.5ex]\draw[#1,radius=#2] (-0.1,0.1) -- (0,-0.1) -- (0.1,0.1);}%
\newcommand{\tikzltriangle}[2][red,fill=red]{\tikz[baseline=-0.5ex]\draw[#1,radius=#2] (-0.1,0) -- (0.1,-0.1) -- (0.1,0.1);}%
\newcommand{\tikzrtriangle}[2][red,fill=red]{\tikz[baseline=-0.5ex]\draw[#1,radius=#2] (0.1,0) -- (-0.1,-0.1) -- (-0.1,0.1);}%
\newcommand{\tikzhexagon}[2][red,fill=red]{\tikz[baseline=-0.5ex]\draw[#1,radius=#2] (0,0.1) -- (0.1,0.05) -- (0.1,-0.05) -- (0,-0.1) -- (-0.1,-0.05) -- (-0.1,0.05);}%
\newcommand{\tikzdiamond}[2][red,fill=red]{\tikz[baseline=-0.5ex]\draw[#1,radius=#2] (0.1,0) -- (0,-0.1) -- (-0.1,0) -- (0,0.1);}%

\tikzset{
  basic/.style  = {draw, text width=2cm, drop shadow, font=\sffamily, rectangle},
  root/.style   = {basic, rounded corners=2pt, thin, align=center,
                   fill=blue!30},
  level 2/.style = {basic, rounded corners=6pt, thin,align=center, fill=blue!30,
                   text width=8em},
  level 3/.style = {basic, thin, align=left, fill=pink!60, text width=6.5em}
}

\title{Survey of Action Recognition, Spotting and Spatio-Temporal Localization in Soccer - Current Trends and Research Perspectives}

\author{ \href{https://orcid.org/0000-0003-0617-7301}{\includegraphics[scale=0.06]{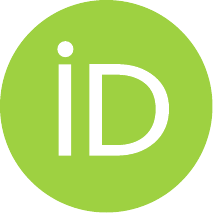}\hspace{1mm}Karolina~Seweryn} \\
	NASK - National Research Institute,\\
	Warsaw, Poland \\
	\texttt{karolina.seweryn@nask.pl} \\
	\And
	\href{https://orcid.org/0000-0002-3407-7570}{\includegraphics[scale=0.06]{orcid.pdf}\hspace{1mm}Anna Wr{\' o}blewska} \\
	Faculty of Mathematics and Information Science,\\
	Warsaw University of Technology,\\
	Warsaw, Poland \\
	\texttt{anna.wroblewska1@pw.edu.pl} \\
    \And
	\href{https://orcid.org/0000-0001-6716-610X}{\includegraphics[scale=0.06]{orcid.pdf}\hspace{1mm}Szymon Łukasik} \\
	Faculty of Physics and Applied Computer Science,\\
	AGH University of Science and Technology,\\
	Cracow, Poland \\
	\texttt{slukasik@agh.edu.pl} \\
}

\renewcommand{\shorttitle}{\textit{arXiv} Template}

\hypersetup{
pdftitle={A template for the arxiv style},
pdfsubject={q-bio.NC, q-bio.QM},
pdfauthor={David S.~Hippocampus, Elias D.~Striatum},
pdfkeywords={First keyword, Second keyword, More},
}

\begin{document}
\maketitle

\begin{abstract}
Action scene understanding in soccer is a challenging task due to the complex and dynamic nature of the game, as well as the interactions between players. This article provides a comprehensive overview of this task divided into action recognition, spotting, and spatio-temporal action localization, with a particular emphasis on the modalities used and multimodal methods. We explore the publicly available data sources and metrics used to evaluate models' performance. The article reviews recent state-of-the-art methods that leverage deep learning techniques and traditional methods. We focus on multimodal methods, which integrate information from multiple sources, such as video and audio data, and also those that represent one source in various ways. The advantages and limitations of methods are discussed, along with their potential for improving the accuracy and robustness of models. Finally, the article highlights some of the open research questions and future directions in the field of soccer action recognition, including the potential for multimodal methods to advance this field. Overall, this survey provides a valuable resource for researchers interested in the field of action scene understanding in soccer.
\end{abstract}

\keywords{action recognition,
   action spotting,
   soccer datasets,
   spatio-temporal action localization,
   modality fusion,
   multimodal learning
}



\maketitle

\section{Introduction}

Soccer is  one of the most popular and lucrative sports worldwide, with billions of fans and many players. In recent years, there has been an increasing interest in using computer vision and machine learning techniques to automatically extract information from match recordings to get valuable insights about the strengths and weaknesses of teams. Understanding the actions that occur during a match is essential for both coaches and players to improve performance and gain a competitive edge. Similarly, scouts visit sports clubs to evaluate the performance and actions of young players to identify those with the most talent that could later be transferred to higher leagues. Automatic retrieval of such information could support scouts' decisions, saving money and time. There are many possible applications
of this process in the television industry. For example, the ability to recognize game actions can enable producers to optimize and automate the broadcast production process, emphasizing key aspects of the game to enhance spectator engagement.  It is particularly valuable for real-time camera selection, post-game studio soccer analytics, and automatic highlights generation.

Action scene understanding 
has become an increasingly important area of research in the context of soccer~\citep{soccernet, soccernet-v2, multisports-li-yixuan}. It poses unique challenges due to the complex and dynamic nature of the game. Players move quickly and often obscure each other, making it difficult to accurately track their movements. Moreover, soccer matches involve a wide range of actions, from simple passes to complex passages of play, tackles and shots on goal, which require different levels of analysis. With recent advancements in machine learning and computer vision, researchers have been exploring various approaches to improve the accuracy of action recognition in soccer. In particular, multimodal methods, which combine data from different sources such as video, audio, and other data, have shown promise in improving the accuracy and robustness of action recognition systems. These methods mirror how humans understand the world by utilizing multiple senses to process data. Employing multiple heterogeneous sources to train models presents both challenges and opportunities. The potential advantage is the improvement of model performance compared to unimodal representation, as incorporating additional modalities provides new information and reveals previously unseen relationships. However, multimodal learning also presents certain difficulties. Information from various sources can be redundant, and this should be filtered out in data representation as one vector. Some solutions build models for all modalities and then create a model combining the predictions. Another approach is to prepare an appropriate joint feature representation. The video data recorded during soccer games often includes information about fans' reactions and audio commentary. Also, Internet websites provide games and player statistics, live comments, and textual data with team analysis. Thus, soccer data can be valuable for researchers experimenting with multimodal learning.

This survey provides a comprehensive overview of recent research on action scene understanding, including action recognition (classification of actions in the trimmed video), spotting (detection and classification of actions in the untrimmed video) and spatio-temporal action localization (classification and tracking of specific actions and objects) in soccer, with a particular focus on available modalities and multimodal approaches. We explore the different data sources used in these methods, the various feature extraction and fusion techniques, and the evaluation metrics used to assess their performance. Also, we discuss the challenges and opportunities in this field, as well as the limitations and future research directions. By analyzing the state-of-the-art methods and identifying their strengths and weaknesses, this survey aims to provide a clear and up-to-date overview of the progress made in action recognition in soccer and to provide insights for researchers in this rapidly evolving area.

The following are the main contributions of this comprehensive literature review:
\begin{itemize}
    \item We define three tasks in the area of soccer action understanding: action recognition, spotting, and spatio-temporal action localization, along with metrics used to assess the performance of these models.
    \item We prepare a list of soccer datasets for action understanding, highlighting the potential of applying multimodal methods.
    \item We examine a variety of existing state-of-the-art models in action recognition, spotting, and spatio-temporal action localization used in soccer as described in the literature.
    \item Based on the thorough assessment and in-depth analysis, we outline a number of key unresolved issues and future research areas that could be helpful for researchers.
\end{itemize}

The article is organized as follows. Subsection~\ref{subsec:motivation} highlights why this survey is important and what distinguishes it from others, while a discussion on the potential of using multimodal sources during training soccer models can be found in Section~\ref{subsec:potential_multimodality}. Section~\ref{sec:research_strategy} describes the research strategy. Tasks related to soccer action scene understanding are described in Section~\ref{sec:problem_description}, and associated metrics and datasets are introduced in Section~\ref{sec:metrics} and Section~\ref{sec:datasets} respectively. Section~\ref{sec:methods} presents methods addressing the three analysed tasks. Future research directions, including the potential for multimodal methods, are discussed in the last section.

\subsection{Motivation}
\label{subsec:motivation}

\begin{table*}
\centering
  \caption{
  List of previous similar reviews. \tikzcircle[green, fill=green]{3pt} denotes articles related to soccer, while \tikzsquare[red, fill=red]{3pt} indicated articles about sport in general. }
  \label{tab:other-surveys}
  \begin{tabular}{p{12cm}c}
    \toprule
    \textbf{Review}  & \textbf{Topic} \\
    \midrule
    A review of vision-based systems for soccer video analysis~\citep{review-soccer-orazio-2010} 
    & \tikzcircle[green, fill=green]{3pt} \\
    Multimodal feature extraction and fusion for semantic mining of soccer video: A survey~\citep{soccer-multimodal-survey-2012} 
    & \tikzcircle[green, fill=green]{3pt} \\
    A survey on event recognition and summarization in football Videos~\citep{survey_soccer_Patil_undated-ke_2014}& 
    \tikzcircle[green, fill=green]{3pt} \\
    Computer vision for sports: current applications and research topics~\citep{survey_sport_graham_2017}
    & \tikzsquare[red, fill=red]{3pt} \\
    A Survey of Video Based Action Recognition in Sports~\citep{survey_sports_2018}  & 
    \tikzsquare[red, fill=red]{3pt} \\
    A comprehensive review of computer vision in sports: open issues, future trends and research directions~\citep{survey_sport_naik_2022} 
    & \tikzsquare[red, fill=red]{3pt} \\
    A survey on video action recognition in sports: datasets, methods and applications~\citep{survey_sport_wu_fei_2022} 
    & \tikzsquare[red, fill=red]{3pt} \\
    Use of deep learning in soccer videos analysis: survey~\citep{survey_sport_akan_2022} 
    & \tikzcircle[green, fill=green]{3pt} \\

  \bottomrule
\end{tabular}
\end{table*}

Several publications listed in Table~\ref{tab:other-surveys} have appeared in recent years reviewing machine learning systems that address the needs of the sports industry. 
Surveys~\citep{survey_sport_graham_2017, survey_sports_2018, survey_sport_naik_2022, survey_sport_wu_fei_2022} show applications of computer vision to automatic analysis of various sports, two of which focus on the action recognition task. However, these surveys do not provide a comprehensive analysis of action detection in soccer, as different sports have different game motions and action types. Therefore, a detailed analysis of dedicated datasets and methods specific to soccer is necessary. 
While there are articles that focus on soccer~\citep{review-soccer-orazio-2010, soccer-multimodal-survey-2012, survey_soccer_Patil_undated-ke_2014, survey_sport_akan_2022}, three of them were published before the release of relevant datasets SoccerNet, SoccerNet-v2, and SoccerNet-v3~\citep{soccernet, soccernet-v2, soccernet-v3}, which caused significant development of this field, including transformer-based solutions. Regarding action recognition, these articles describe binary models that classify only certain actions, such as goals or offside. 
A comprehensive review of various tasks was published in article~\citep{survey_sport_wu_fei_2022} in 2022. Only one reported action recognition solution is evaluated on SoccerNet-v2~\citep{soccernet-v2}, more specifically benchmark result is reported. Also, spatio-temporal action localization is not described in this publication. The potential of using multimodal inputs is only briefly described in previous surveys. Only one publication~\citep{soccer-multimodal-survey-2012} addresses this topic; however, the methods outlined therein are not considered state-of-the-art nowadays. Mentioned publications focus mainly on action classification (recognition), while action localization in time or spatio-temporal space is more relevant in real-life scenarios.




\subsection{Potential of Using Multimodality}
\label{subsec:potential_multimodality}

Soccer is a sport that generates a vast amount of data, including videos, information on players, teams, matches, and events. Numerous matches are documented, and these recordings provide significant insights into the game. They vary in terms of video quality, used device and camera perspective (drone, single-camera, multiple cameras). Beyond the raw video feed, many characteristics can be extracted from video streams, including optical flow, players and ball trajectories, players' coordinates. When audio is available, it could be an additional data source  capturing people's reactions, which can be represented using techniques like Mel Spectrogram~\citep{xu2005hmm}. Also, commentary data can be transcribed using automatic speech recognition systems such as Whisper~\citep{radford2022whisper} or vosk\footnote{\url{https://alphacephei.com/vosk/}}. Unstructured texts containing match reports can be scrapped from various websites. Soccer clubs gather and analyse a vast amount of information about players to gain an advantage and choose the optimal tactics. Data from GPS systems, accelerometers, gyroscopes, or magnetometers can also be useful in soccer data analysis.

Combining many modalities is proven to achieve better results than using unimodal representations~\citep{neurips-multimodal-fusion-nagrani-arsha}. Soccer match recordings already contain many data types with predictive potential: audio, video, and textual transcriptions. Additional information about the outcomes of games or the time of individual events is also regularly reported on many websites. To sum up, investigating multimodal approaches is a logical step in soccer data analysis due to the accessibility of diverse data sources. As far as we know, there is no survey on the application of modality fusion in action recognition, spotting, and spatio-temporal action localization in sports videos. This survey provides a comprehensive overview of the current state of research and helps advance the field of action understanding in soccer by identifying areas for future research.






\section{Definition of Research Strategy}
\label{sec:research_strategy}

The articles published between 2000 and 2022 are included in this survey. In order to find related articles, we used online databases such as Scopus~\footnote{\url{https://www.scopus.com/}}, ScienceDirect~\footnote{\url{https://www.sciencedirect.com/}}, ACM~\footnote{\url{https://dl.acm.org/}}, IEEE Xplore~\footnote{\url{https://ieeexplore.ieee.org/Xplore/home.jsp}}, SpringerLink~\footnote{\url{https://link.springer.com/}} and SemanticScholar~\footnote{\url{https://www.semanticscholar.org/}} and keyword search. The primary search keys were: \textit{multimodal, multimodality, action recognition, sport, activity recognition, event recognition, event classification, action spotting, event detection, modality fusion, video, audio, text, activity localization, spatio-temporal action recognition, football, soccer, action localization, 
soccer dataset, soccernet}. Each query produced several articles. Additionally, we manually added many papers to our list by analyzing the references of the papers we identified. According to their relevance and year of publication, some of them were excluded from this analysis. 

\section{Problem Description}
\label{sec:problem_description}

\subsection{Actions}
\label{sec:action}



\begin{figure}[h]
    \centering
    \includegraphics[width=0.3\textwidth]{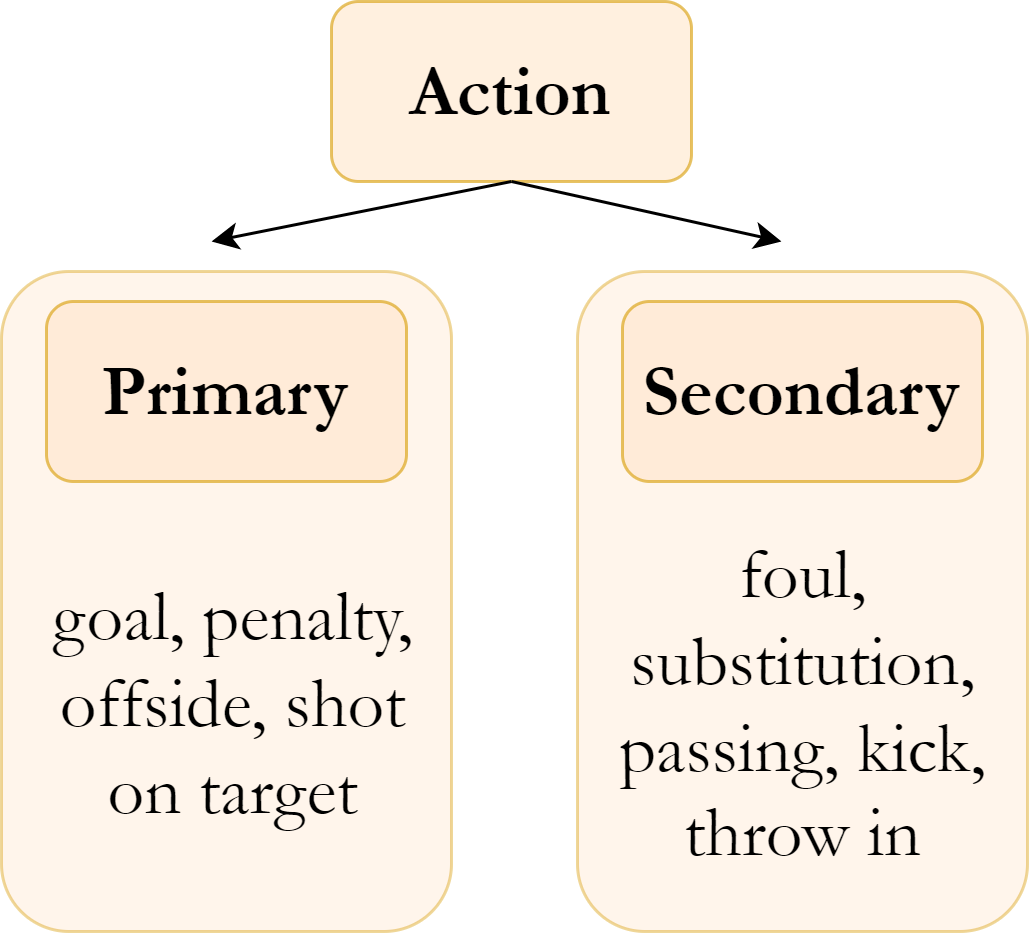}
    \caption{Difference between primary and secondary actions.}
    \label{fig:action_types_primary_secondary}
\end{figure}

According to \citep{soccer-multimodal-survey-2012}, soccer actions can be divided based on their relevance into primary and secondary events as depicted in Figure~\ref{fig:action_types_primary_secondary}. The primary events directly affect the match's outcome and can directly cause goal opportunities, while the secondary actions are less important and do not affect the result of the match as much.

\begin{figure*}[h]
    \centering
    \includegraphics[width=\textwidth]{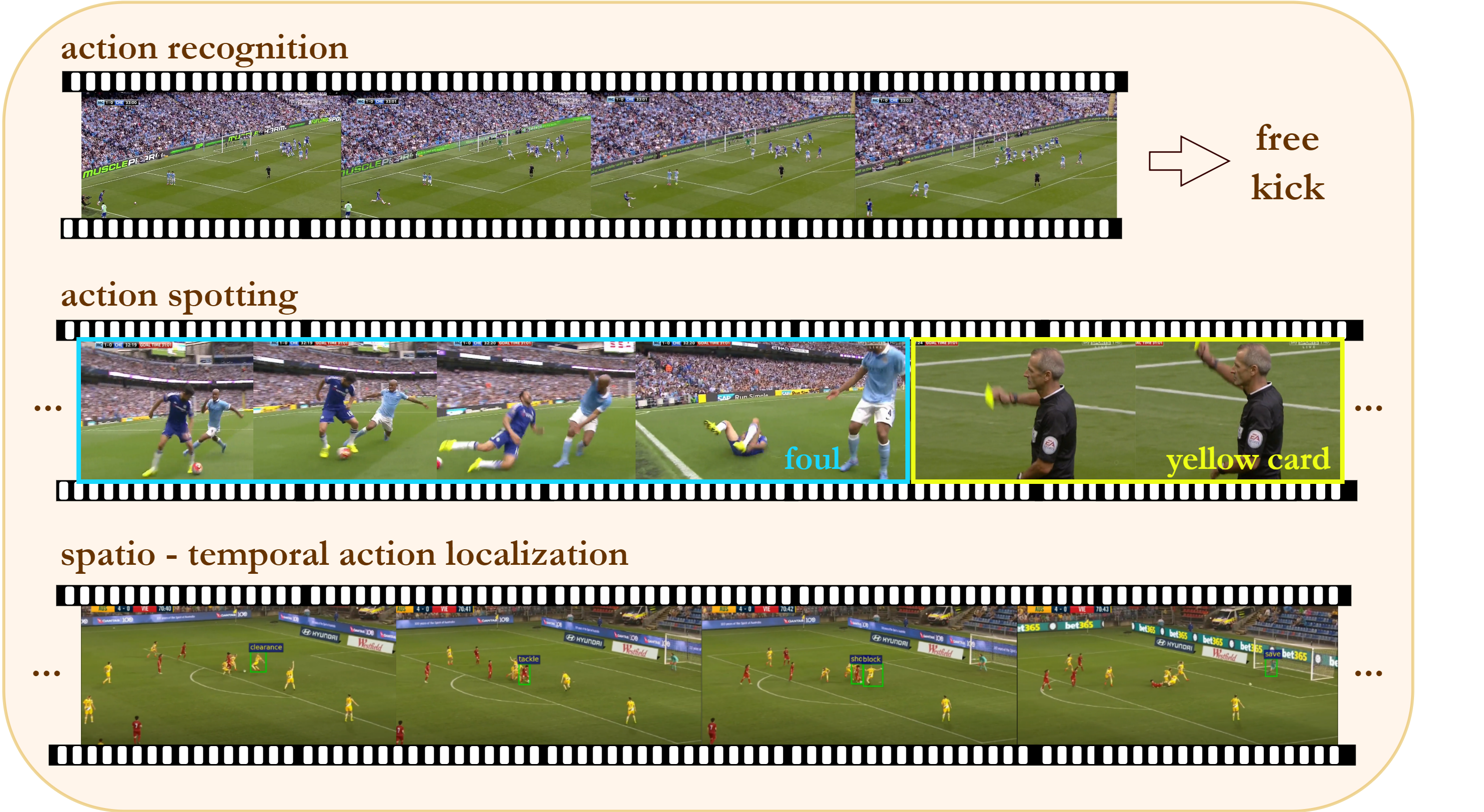}
    \caption{Comparison of tasks related to action analysis. Frames used in this visualization are from SoccerNet~\citep{soccernet-v2} and MultiSports~\citep{multisports-li-yixuan} datasets.} 
    \label{fig:action-recognition-types}
\end{figure*}

Action analysis can also be divided into three tasks that differ in their use of information about the time and localization of the event: action recognition, action spotting, and spatio-temporal action detection. Figure~\ref{fig:action-recognition-types} highlights differences between mentioned tasks. 

\textbf{Action recognition}, also known as \textbf{action/event classification}, is the classification of an event in a trimmed video. The model receives as input a series of frames; for the entire series, it has to predict which class the video refers to. In contrast, action spotting is a slightly different task which involves identifying the segment of the untrimmed video in which the action occurs, and then classifying it into predefined categories. An action can be temporally localized by defining a boundary that contains a start and end timestamp or a single frame timestamp, such as the start of the action. \textbf{Action spotting} is also referred to as \textbf{temporal action detection} or \textbf{temporal action localization}. In addition, we can distinguish an extension of this problem that incorporates actor location data called \textbf{spatio-temporal action detection (localization)}. This task aims to detect temporal and spatial information about action as moving bounding boxes of players. This involves detecting the location, timing, and type of actions performed by players, such as passing, shooting, tackling, dribbling, and goalkeeping. This task is particularly relevant in the analysis of an individual’s behaviour and performance.




\subsection{Multimodality}
\label{sec:multimodality}

\begin{figure}[h]
    \centering
    \includegraphics[width=0.42\textwidth]{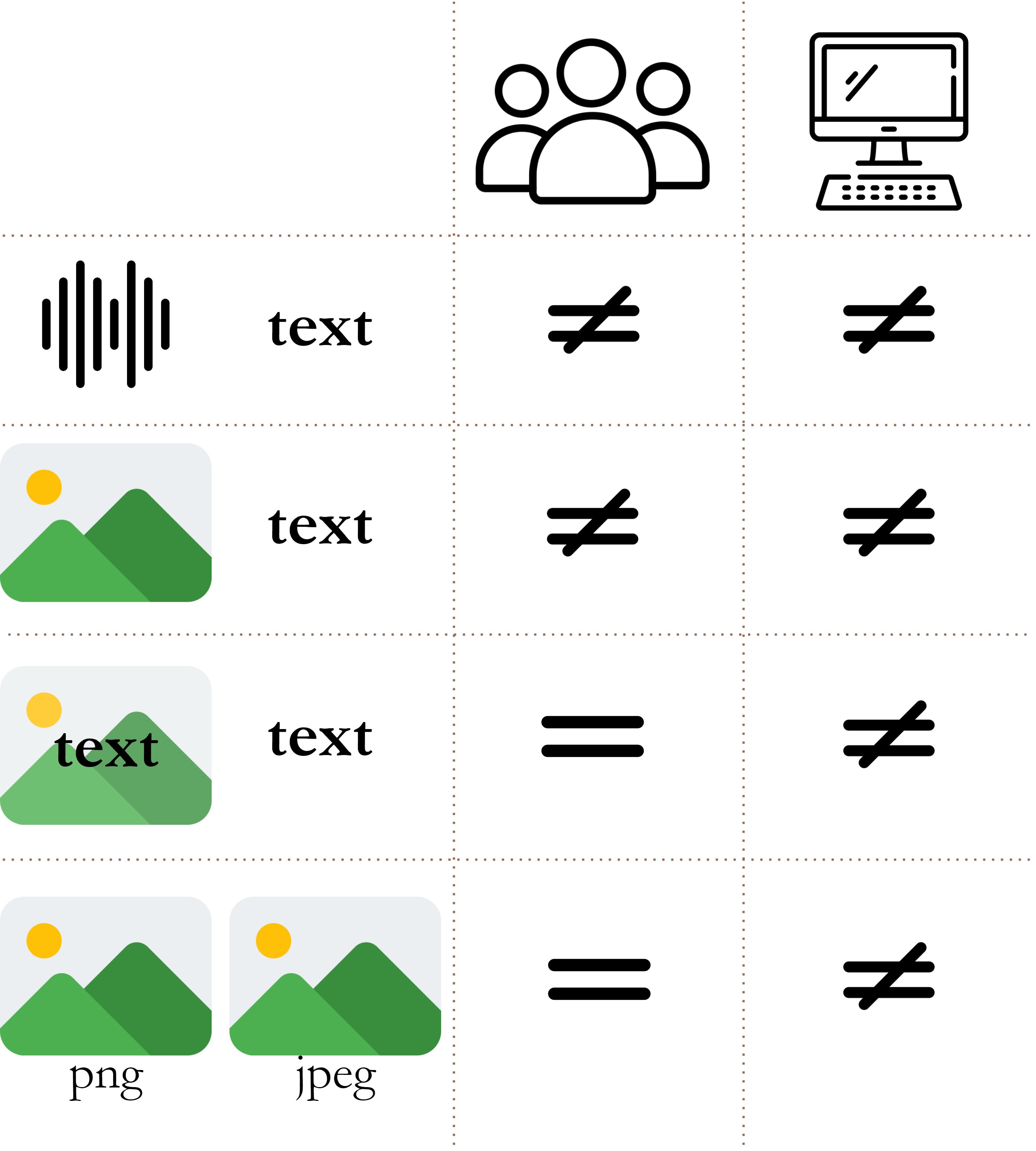}
    \caption{Difference between extracting information by people and machines. = denotes the same data, while $\neq$ means different. Visualization inspired by~\citet{parcalabescu2021multimodality}.}
    \label{fig:is_it_multimodality}
\end{figure}

Although we intuitively know what multimodality is, the formal definitions differ. One perspective is more human-centred and refers to the way people perceive the world, and other definitions are more related to machine-readable representations.
According to~\citet{baltruvsaitis2018multimodal}, \textit{"Our experience of the world is multimodal -- we see objects, hear sounds, feel the texture, smell odours, and taste flavours. Modality refers to the way in which something happens or is experienced and a research problem is characterized as multimodal when it includes multiple such modalities."} (definition 1: human-centered) while \citet{guo2019deep_multimodal_representation_survey} define modality as \textit{"a particular way or mechanism of encoding information"} (definition 2: machine-centered). Therefore, different encodings for one source (e.g. BERT embeddings and TFIDF encoding for text) would be considered multimodal according to definition 2, but not according to definition 1. This inconsistency was noted in~\citep{parcalabescu2021multimodality}, where authors proposed a new task-relative definition of multimodality: \textit{"A machine learning task is multimodal when inputs or outputs are represented differently or are composed of distinct types of atomic units of information"}. Figure~\ref{fig:is_it_multimodality} highlights that inputs can be processed differently by people and machines. For instance, while people may perceive identical meaning in textual content presented in either text or image form, machine-extracted data can indicate significant dissimilarities. In this work, we analyse both human-centred and machine-centred multimodal models.



\section{Metrics}
\label{sec:metrics}

The most common metric measuring classification performance among all analysed articles is mAP (mean average precision). In the case of action spotting avg-mAP is used, and for spatio-temporal action detection, Video-mAP@$\delta$ and Frame-mAP@$\delta$ are used. This section explains and summarizes various metrics used to measure the accuracy of the model's predictions.

\subsection{Action recognition}

\textbf{Precision} measures the model's ability to detect only relevant actions in a video. It is defined as the fraction of correctly classified samples (TP) out of all positive predictions (TP + FP):

$$\varB{Precision} = \frac{TP}{TP + FP}.$$

\textbf{Recall} describes how well the model detects all relevant samples. It is defined as the fraction of correctly classified samples (TP) out of all positive ground truth (TP + FN):

$$\varB{Recall} = \frac{TP}{TP + FN}.$$

\textbf{F1-score} is the harmonic mean of recall and precision:

$$\varB{F1-score} = 2 \cdot \frac{precision \cdot recall}{precision + recall}.$$

The objective of the model is to achieve both high precision and recall. However, in practice, a trade-off between these metrics is chosen. The relationship between them can be depicted through \textbf{Precision-Recall (PR) curve}, which illustrates the values of precision on the y-axis and recall on the x-axis for different thresholds of model's confidence score. 
The term \textbf{AP} is an abbreviation for average precision and refers to the area under the precision-recall curve computed across all recall values. It is worth mentioning that a high AP value indicates a balance of both high precision and recall. Given that the PR curve often exhibits a zigzag pattern, calculating the area under the curve (AUC) accurately can be challenging. Therefore, various interpolation techniques, such as  11-points interpolation~\citep{Zhang2009_11p_interp_prec_recall} or interpolation based on all points, are commonly employed. 11-points precision-recall curve interpolates precision at 11 recall levels (0.0, 0.1, \dots, 1.0). The interpolation of precision ($P_{interp}(R)$ at recall level $R$ is defined as the maximum precision with a recall value greater or equal than $R$ and is defined in the equation below.

$$P_{interp}(R) = max_{R' \geq R} P(R').$$

Thus, the estimation of AP value can be defined as the mean of interpolated precisions over 11 recall values:

$$AP_{11} = \frac{1}{11} \sum_{R \in \{0.0, 0.1, \dots, 1.0\}} P_{interp}(R)$$

Interpolation based on all points takes all points into account. AP is computed as the weighted mean of precisions at each recall (R) level. The weights are differences between current and previous recall values.

$$AP_{all} = \sum_{n=1}^{N} (R_{n} - R_{n-1}) P_{interp}(R),$$

where 

$$P_{interp}(R_{n}) = max_{R' \geq R_{n}} P(R').$$

\textbf{mean AP (mAP)} measures the accuracy of a classifier over all classes and is defined as 
$$\varB{mAP} = \frac{1}{|\Gamma|} \sum_{\gamma \in \Gamma}^{} \varB{AP(\gamma)},$$

where $\Gamma$  is a set of all possible classes and $AP(\gamma)$ denotes AP of class $\gamma$.

\textbf{Top-N Accuracy} is a fraction of correct predictions, where prediction is assigned as correct when any of the model's N highest probability scores match the expected answer (ground truth). In other words, it measures the proportion of correctly classified instances where the predicted class is among the N most probable classes. Top-N accuracy is a useful metric in situations where the exact class prediction may not be critical, but rather the identification of a set of probable classes that may be relevant.

\subsection{Action spotting}

Unlike the classic classification, action spotting also considers the time in which the action takes place. This aspect is also reflected in the metrics used to assess the models.

\textbf{avg-mAP} is the average of mAP for different tolerance values and can be defined as

$$\varB{avg-mAP} = \frac{1}{|\Theta|} \sum_{\theta \in \Theta}^{} \varB{mAP}(\theta),$$

where $\Theta$  is a set of different tolerances. Authors of \citep{soccernet-v2} divided this metric into two subgroups: \textbf{loose avg - mAP} for tolerances $5$s, $10$s,\dots, $60$s, and tight-avg-mAP  with tolerance $1$s, $2$s, \dots, $5$s. If the high accuracy of action localization is required then \textbf{tight-avg-mAP} will be more appropriate.

\subsection{Spatio-temporal action detection}

Not only the time aspect but also the localization (bounding box coordinates) is important in spatio-temporal action detection. This section begins by introducing an assessment of the accuracy of the location of the action in a frame and then describes the associated metrics.

\textbf{Intersaction over Union (IoU)}, also called Jaccard Index, measures the overlap of the predicted bounding box and the ground truth bounding box. IoU is invariant to the scale which means that the similarity between two shapes is not affected by the scale of their space.

$$IoU = \frac{\text{Area of Overlap}}{\text{Area of Union}}= \frac{P \cap GT}{P \cup GT} =
\frac{
    \tikz{\fill[draw=blue, very thick, fill=blue!30] (0,0) rectangle (1,1) (0.25,-0.25) rectangle (1.25,0.75);
    \fill[draw=blue, very thick, fill=white, even odd rule] (0,0) rectangle (1,1) (0.25,-0.25) rectangle (1.25,0.75);}}
{\tikz{\fill[draw=blue, fill=blue!30, very thick] (0,0) rectangle (1,1) (0.25,-0.25) rectangle (1.25,0.75);}},$$

where P denotes prediction and GT means ground truth. An action is considered correctly classified and localized if the IoU between the predicted and the ground-truth bounding boxes is above a threshold  $\delta$.

\paragraph{3D Intersection over Union (3D IoU)}

The intersection over union in 3D is computed as an overlap between two cuboids. 

$$3D\; IoU = \frac{\text{Volume of Overlap}}{\text{Volume of Union}}= \frac{P \cap GT}{P \cup GT} = $$
$$ =
 \frac{
\begin{tikzpicture}
\pgfmathsetmacro{\cubedx}{0.3}
\pgfmathsetmacro{\cubedy}{0.3}
\pgfmathsetmacro{\cubekx}{0.5}
\pgfmathsetmacro{\cubeky}{0.5}
\pgfmathsetmacro{\cubex}{2}
\pgfmathsetmacro{\cubey}{1}
\pgfmathsetmacro{\cubez}{1}

\coordinate (A1) at (\cubedx,0,0);
\coordinate (A2) at (\cubex+\cubedx,0,0);
\coordinate (A3) at (\cubex+\cubedx,\cubey,0);
\coordinate (A4) at (\cubedx,\cubey,0);
\coordinate (A5) at (\cubedx,0,-\cubez);
\coordinate (A6) at (\cubex+\cubedx,0,-\cubez);
\coordinate (A7) at (\cubex+\cubedx,\cubey,-\cubez);
\coordinate (A8) at (\cubedx,\cubey,-\cubez);

\coordinate (B1) at (0,\cubedy,0);
\coordinate (B2) at (\cubex,\cubedy,0);
\coordinate (B3) at (\cubex,\cubey+\cubedy,0);
\coordinate (B4) at (0,\cubey+\cubedy,0);
\coordinate (B5) at (0,\cubedy,-\cubez);
\coordinate (B6) at (\cubex,\cubedy,-\cubez);
\coordinate (B7) at (\cubex,\cubey+\cubedy,-\cubez);
\coordinate (B8) at (0,\cubey+\cubedy,-\cubez);

\coordinate (C1) at (\cubex,\cubey,0);
\coordinate (C2) at (\cubex,\cubey,-\cubez);
\coordinate (C3) at (\cubedx,\cubedy,0);
\coordinate (C4) at (\cubedx,\cubedy,-\cubez);

\draw[draw=none,fill=blue!30] (A4) -- (A8) -- (C2) -- (C1) -- (A4);
\draw[draw=blue, very thick, dashed] (A4) -- (A8) -- (C2) -- (C1);
\draw[draw=none, fill=blue!30] (A4) -- (C3) -- (B2) -- (C1) -- (A4);
\draw[draw=none, fill=blue!30] (C1) -- (C2) -- (B6) -- (B2) -- (C1);
\draw[draw=blue, very thick] (A1) -- (A2) -- (A3) -- (A4) -- (A1);
\draw[draw=blue, very thick] (A6) -- (A7) -- (C2);
\draw[draw=blue, very thick, dashed] (A8) -- (A5) -- (A6);
\draw[draw=blue, very thick] (B1) -- (B2) -- (B3) -- (B4) -- (B1);
\draw[draw=blue, very thick] (B6) -- (B7) -- (B8);
\draw[draw=blue, very thick, dashed] (B8) -- (B5) -- (B6);
\draw[draw=blue, very thick, dashed] (B5) -- (B1);
\draw[draw=blue, very thick] (B4) -- (B8) -- (B7) -- (B3);
\draw[draw=blue, very thick] (A2) -- (A6);
\draw[draw=blue, very thick] (A3) -- (A7);
\draw[draw=blue, very thick, dashed] (A5) -- (A1);
\draw[draw=blue, very thick, dashed] (B2) -- (B6);
\end{tikzpicture}
}{
\begin{tikzpicture}
\pgfmathsetmacro{\cubedx}{0.3}
\pgfmathsetmacro{\cubedy}{0.3}
\pgfmathsetmacro{\cubekx}{0.5}
\pgfmathsetmacro{\cubeky}{0.5}
\pgfmathsetmacro{\cubex}{2}
\pgfmathsetmacro{\cubey}{1}
\pgfmathsetmacro{\cubez}{1}

\coordinate (A1) at (\cubedx,0,0);
\coordinate (A2) at (\cubex+\cubedx,0,0);
\coordinate (A3) at (\cubex+\cubedx,\cubey,0);
\coordinate (A4) at (\cubedx,\cubey,0);
\coordinate (A5) at (\cubedx,0,-\cubez);
\coordinate (A6) at (\cubex+\cubedx,0,-\cubez);
\coordinate (A7) at (\cubex+\cubedx,\cubey,-\cubez);
\coordinate (A8) at (\cubedx,\cubey,-\cubez);

\coordinate (B1) at (0,\cubedy,0);
\coordinate (B2) at (\cubex,\cubedy,0);
\coordinate (B3) at (\cubex,\cubey+\cubedy,0);
\coordinate (B4) at (0,\cubey+\cubedy,0);
\coordinate (B5) at (0,\cubedy,-\cubez);
\coordinate (B6) at (\cubex,\cubedy,-\cubez);
\coordinate (B7) at (\cubex,\cubey+\cubedy,-\cubez);
\coordinate (B8) at (0,\cubey+\cubedy,-\cubez);

\coordinate (C1) at (\cubex,\cubey,0);
\coordinate (C2) at (\cubex,\cubey,-\cubez);
\coordinate (C3) at (\cubedx,\cubedy,0);
\coordinate (C4) at (\cubedx,\cubedy,-\cubez);

\draw[fill=blue!30,draw=none] (A1) -- (A2) -- (A3) -- (A4) -- (A1);
\draw[fill=blue!30,draw=none] (A5) -- (A6) -- (A7) -- (A8) -- (A5);
\draw[fill=blue!30,draw=none] (B1) -- (B2) -- (B3) -- (B4) -- (B1);
\draw[fill=blue!30,draw=none] (B5) -- (B6) -- (B7) -- (B8) -- (B5);
\draw[fill=blue!30,draw=none] (A2) -- (A6) -- (A7) -- (A3) -- (A2);
\draw[fill=blue!30,draw=none] (B1) -- (B4) -- (B8) -- (B5) -- (B1);
\draw[fill=blue!30,draw=none] (A4) -- (A8) -- (C2) -- (C1) -- (A4);
\draw[draw=blue, very thick, dashed] (A4) -- (A8) -- (C2) -- (C1);
\draw[draw=blue, very thick] (A1) -- (A2) -- (A3) -- (A4) -- (A1);
\draw[draw=blue, very thick] (A6) -- (A7) -- (C2);
\draw[draw=blue, very thick, dashed] (A8) -- (A5) -- (A6);
\draw[draw=blue, very thick] (B1) -- (B2) -- (B3) -- (B4) -- (B1);
\draw[draw=blue, very thick] (B6) -- (B7) -- (B8);
\draw[draw=blue, very thick, dashed] (B8) -- (B5) -- (B6);
\draw[draw=blue, very thick, dashed] (B5) -- (B1);
\draw[draw=blue, very thick] (B4) -- (B8) -- (B7) -- (B3);
\draw[draw=blue, very thick] (A2) -- (A6);
\draw[draw=blue, very thick] (A3) -- (A7);
\draw[draw=blue, very thick, dashed] (A5) -- (A1);
\draw[draw=blue, very thick, dashed] (B2) -- (B6);
\end{tikzpicture}
}
$$

In the study introducing MultiSports dataset~\citep{multisports-li-yixuan} they defined 3D IoU (spatio-temporal-IoU) as IoU over the temporal domain multiplied by the average of the IoU between the overlapped frames (also used in~\citep{Singh2022SpatioTemporalAD, Weinzaepfel2015LearningTT}).



\paragraph{MABO (Mean Average Best Overlap)}

\textbf{ABO} (Average Best Overlap)~\citep{mabo-Uijlings2013SelectiveSF,Kalogeiton2017ActionTD} for class $c$ is defined as

$$ABO(c) = \frac{1}{|G^{c}|} \sum_{g_{i}^{c} \in G^{c}}  max_{l_j \in L } IoU(g_{i}^{c}, l_{j}),$$

\noindent where $G^{c}$ is a set of all ground truths for class $c$ and $L$ is predicted bounding box. The intersection over union (IoU) between every ground truth and predicted boxes (or tubes) is computed. Then, for each ground-truth box or tube, the overlap of the detection with the highest IoU value is retained (best-overlapping detection BO). Next, for every class, an average of all maximum intersections is calculated. The mean of \textbf{ABO} over all classes is called \textbf{MABO}.

\paragraph{Video-mAP@$\delta$ and Frame-mAP@$\delta$}

Two groups of metrics can be considered to evaluate spatio-temporal action detectors: frame and video level. In the case of frame-level, metrics such as AP are computed for defined IoU threshold $\delta$. Prediction is considered correct if its IoU with a ground truth box is greater than a given threshold and the predicted label matches the ground truth one. Then, mean AP is computed by averaging over all classes. In the literature, it is often referred to as \textbf{frame-mAP@$\delta$}, \textbf{f@$\delta$} or \textbf{f-mAP@$\delta$}. 3D IoU between predicted tubes and ground truth tubes is often used to report video-level metrics. A model returns the correct tube if its 3D IoU with ground truth tube is above $\delta$ and correctly classified. Similarly, averaging the metrics (e.g. AP) over classes gives an overall metric such as \textbf{video-mAP@$\delta$} (also denoted as \textbf{v-mAP@$\delta$} and \textbf{v@$\delta$}). By analogy, \textbf{Precision@{k}} and \textbf{Recall@{k}} can be defined.

\paragraph{Motion mAP and Motion AP}

In article~\citep{Singh2022SpatioTemporalAD}, new metrics considering motion have been introduced. Actions are divided into  three categories based on their motion size (large, medium, and small). With these labels, Average Precision (AP) is computed for each motion category. Computing the AP for each action class and then averaging the results for motion categories is referred to as the \textbf{Motion-mAP} while computing the AP for the motion categories regardless of action class is called the \textbf{MotionAP}. They are calculated at video and frame levels.

\section{Datasets}
\label{sec:datasets}

The potential to use machine learning to analyse tactics and statistics in sports has automatically resulted in a significant increase in publicly available datasets~\citep{soccernet, soccernet-v2, football-actions-tsunoda, karimi2021soccer, multisports-li-yixuan, soccer-logs-data-pappalardo}. 
Television broadcasters record sports games along with commentary, while online platforms offer detailed information and player statistics pertaining to the game.
Thus, creating a multi-modal database should be relatively easy and intuitive.
Table~\ref{tab:datasets-modalities} examines the availability of different modalities in published soccer datasets for action recognition, spotting and spatio-temporal action localization. 

\begin{table*}
\centering
  \caption{Modalities in soccer datasets with annotations of action detection task. Circles mean different tasks: \tikzcircle[green, fill=green]{3pt} - action classification, \tikzsquare[red, fill=red]{3pt} - action spotting, \tikzltriangle[teal, fill=teal]{3pt} - spatio-temporal action localization, \tikzrtriangle[blue, fill=blue]{3pt} - tracking, \tikzinvtriangle[yellow, fill=yellow]{3pt} - camera shot segmentation,  \tikzdiamond[gray, fill=gray]{3pt} - replay grounding, \tikzhexagon[pink, fill=pink]{3pt} - highlight detection, \tikztriangle[orange, fill=orange]{3pt} - object detection. 
  } 
  \label{tab:datasets-modalities}
  \begin{tabular}{p{0.7cm}p{5.8cm}p{1.5cm}p{1cm}p{1cm}p{1.9cm}} 
    \toprule
    & \textbf{Dataset}  & \textbf{Task} & \textbf{Video/ Photo} & \textbf{Audio} &   \textbf{Publicly Available}\\
    \midrule
    
     \multirow{10}{*}{\rotatebox[origin=c]{90}{soccer}}  & SoccerNet~\citep{soccernet}  & \tikzcircle[green, fill=green]{3pt} \tikzsquare[red, fill=red]{3pt} &\ding{51} & \ding{51}  & Yes\\
    
    & SoccerNet-v2~\citep{soccernet-v2} & \tikzcircle[green, fill=green]{3pt} \tikzsquare[red, fill=red]{3pt} \tikzinvtriangle[yellow, fill=yellow]{3pt} \tikzdiamond[gray, fill=gray]{3pt} &\ding{51} & \ding{51}  & Yes\\
    & SoccerNet-v3~\citep{soccernet-v3} & \tikzcircle[green, fill=green]{3pt} \tikzsquare[red, fill=red]{3pt} \tikzinvtriangle[yellow, fill=yellow]{3pt} \tikzdiamond[gray, fill=gray]{3pt} \tikztriangle[orange, fill=orange]{3pt} &\ding{51} & \ding{51}  & Yes\\
    
    
    & Football Action~\citep{football-actions-tsunoda} & \tikzcircle[green, fill=green]{3pt} \tikzrtriangle[blue, fill=blue]{3pt} & \ding{51} & ? &  No\\
    
    & Comprehensive Soccer~\citep{comprehensive-soccer-junqing}  & \tikzcircle[green, fill=green]{3pt} \tikzsquare[red, fill=red]{3pt} \tikzinvtriangle[yellow, fill=yellow]{3pt} \tikzrtriangle[blue, fill=blue]{3pt}  & \ding{51} & \ding{55} & Yes\\
    
    & SSET~\citep{sset-na-feng} & \tikzcircle[green, fill=green]{3pt} \tikzsquare[red, fill=red]{3pt} \tikzinvtriangle[yellow, fill=yellow]{3pt} \tikzrtriangle[blue, fill=blue]{3pt}& \ding{51} & \ding{55} & Yes\\
    
    & SoccerDB~\citep{soccerdb-yudong} & \tikzcircle[green, fill=green]{3pt} \tikzsquare[red, fill=red]{3pt} \tikztriangle[orange, fill=orange]{3pt} \tikzhexagon[pink, fill=pink]{3pt} & \ding{51} & \ding{55} & Yes\\
    
    & Soccer-logs~\citep{soccer-logs-data-pappalardo} & \tikzcircle[green, fill=green]{3pt} \tikzsquare[red, fill=red]{3pt} & \ding{55} & \ding{55} & Yes \\
    & VisAudSoccer~\citep{9106051} & \tikzcircle[green, fill=green]{3pt} \tikzdiamond[gray, fill=gray]{3pt} & \ding{51} & \ding{51} & No \\
    & SEV~\citep{karimi2021soccer} & \tikzcircle[green, fill=green]{3pt} & \ding{51} & \ding{55} & Yes \\
    & EIGD-S~\citep{10.1145/3475722.3482792}  & \tikzcircle[green, fill=green]{3pt} \tikzsquare[red, fill=red]{3pt}  & \ding{51} & \ding{51} & Yes \\
    
    \midrule
    
    
    
    


    
    
    multi sports& MultiSports~\citep{multisports-li-yixuan}   & \tikzltriangle[teal, fill=teal]{3pt} 
    \tikzrtriangle[blue, fill=blue]{3pt}  & \ding{51} & \ding{55} & Yes \\
  \bottomrule
\end{tabular}
\end{table*}

\subsection{Soccer Datasets}
\paragraph{SoccerNet} SoccerNet\footnote{\url{https://www.soccer-net.org/data}}~\citep{soccernet} was introduced as a benchmark dataset for action spotting in soccer. The dataset consists of 500 soccer matches from the main European Championships (764 hours in total) and annotations of three action types: goal, card, and substitution. Each match is divided into two videos: one for each half of the match. The matches are split into a train (300 observations), test (100 observations), and validation datasets (100 observations). Also, the authors published extra 50 observations without labels for the challenge task.

\begin{figure*}
\begin{center}
    \includegraphics[width=0.8\textwidth]{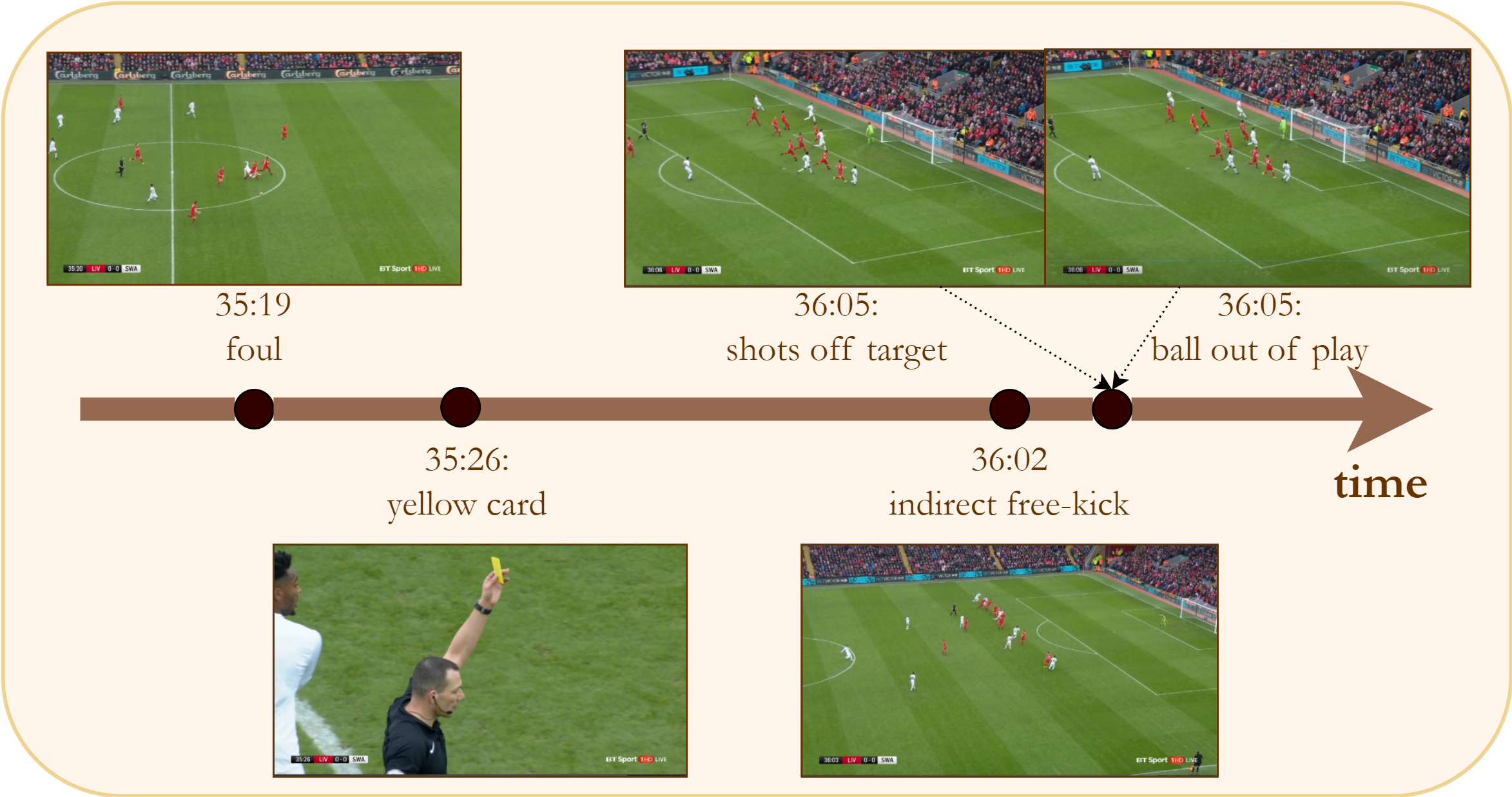}
    \caption{Examples of actions from SoccerNet-v2 dataset~\citep{soccernet-v2}. Frames come from the match between Liverpool and Swansea (2017-01-21 - 15:30).}
    \label{fig:soccernet_dataset_ex}
\end{center}
\end{figure*}

\paragraph{SoccerNet-v2} SoccerNet-v2~\citep{soccernet-v2} enriched the SoccerNet dataset by manually annotating 17 action categories. In contrast to its predecessor, actions occur around 17 times more frequently (8.7 events per hour in SoccerNet, 144 actions per hour in SoccerNet-v2). Table~\ref{tab:soccernet-v2-action-count} shows action types and their frequency. Each action has an assigned visibility category (shown and unshown) to indicate whether an action was shown in the broadcast video. Detecting unshown actions is very difficult and requires understanding the temporal context of actions. Figure~\ref{fig:soccernet_dataset_ex} presents examples of actions for a selected match. Moreover, this dataset includes manual annotations for camera shot segmentation with boundary detection and replay grounding task. It is worth mentioning that SoccerNet-v2 recordings include audio commentary. 


\begin{table}
  \caption{Number of actions in SoccerNet-v2 dataset~\citep{soccernet-v2}.}
  \label{tab:soccernet-v2-action-count}
  \begin{center}
  \begin{tabular}{ccccc}
    \toprule
    \textbf{Action} & \textbf{\#Train} & \textbf{\#Test} & \textbf{\#Valid} & \textbf{Total} \\
    \midrule
     Ball out of play & 19097	&6460	&6253 & 31810\\
     Clearance &  4749 &	1631 & 1516 & 7896\\
     Corner	&2884	&999	&953 & 4836\\
     Direct free-kick&	1379&	382&	439 & 2200 \\
     Foul&	7084&	2414&	2176 & 11674\\
    Goal&	995	&337	&371 & 1703\\
    Indirect free-kick&	6331&	2283&	1907 & 10521\\
    Kick-off	&1516&	514	&536 & 2566\\
    Offside	&1265&	416&	417 & 2098 \\
    Penalty	&96	&41&	36 & 173\\
    Red card	&34&	8	&13 & 55\\
    Shots off target&	3214&	1058	&984 & 5256\\
    Shots on target	&3463&	1175&	1182 & 5820\\
    Substitution	&1700&	579&	560 & 2839\\
    Throw-in&	11391&	3809&	3718 & 18918\\
    Yellow card	&1238&	431	&378 & 2047\\
    Yellow-$>$red card &	24	& 14	&8 & 46\\
  \bottomrule
\end{tabular}
  \end{center}
\end{table}

\begin{figure}[h]
    \centering
    \includegraphics[width=0.5\textwidth]{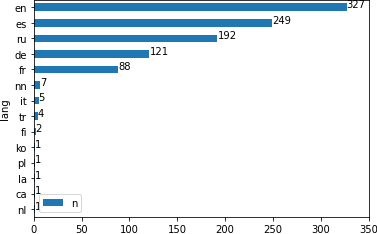}
    \caption{Distribution of commentary languages detected by Whisper~\citep{radford2022whisper} in SoccerNet-v2~\citep{soccernet-v2} dataset. }
    \label{fig:soccernet-v2-lang}
\end{figure}

We analysed audio commentary in the SoccerNet-v2 with ASR (Automatic Speech Recognition) model called whisper~\citep{radford2022whisper}. We noticed that $37$ out of $1000$ observations ($3.7$\%) do not have audio commentary. Figure~\ref{fig:soccernet-v2-lang} shows the distribution of detected languages. Each half of the match is analyzed as a separate observation. Sometimes, the first half of the game's commentary is in a different language than the second.

\paragraph{SoccerNet-v3} SoccerNet-v3~\citep{soccernet-v3} is an extension of SoccerNet-v2~\citep{soccernet-v2} containing spatial annotations and associations between different view perspectives. Action annotations have been enriched with associated frames from the replay clips ($21,222$ of instances have been added). Therefore, it enables the exploration of multi-view action analysis. Also, they added lines and goals annotations, bounding boxes of players and referees ($344,660$ of instances), bounding boxes of objects including ball, flag, red/yellow card ($26,939$ of instances), multi-view player correspondences ($172,622$ of instances) and jersey numbers ($106,592$ of instances).

\begin{figure*}[h]
    \centering
    \includegraphics[width=\textwidth]{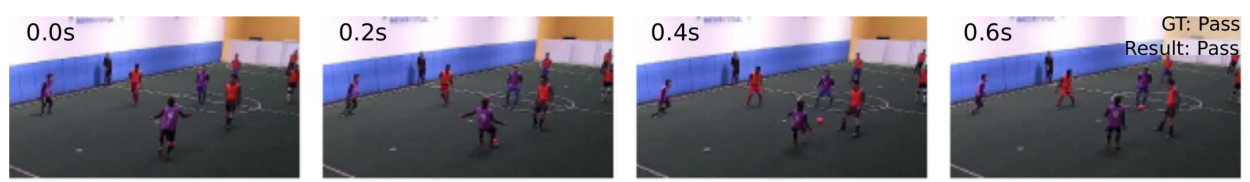}
    \caption{Example from Football actions dataset. Source: \citep{football-actions-tsunoda}.}
    \label{fig:football-action-lstm-dataset-ex1}
\end{figure*}

\paragraph{Football actions} The dataset~\citep{football-actions-tsunoda} consists of two match recordings (each lasting 10 minutes) from 14 cameras located in different places. Five actions are manually annotated: pass, dribble, shoot, clearance, and loose ball. Additionally, the authors provided annotations of the ball and players' 3D positions. Examples of images from this dataset are presented in Figure~\ref{fig:football-action-lstm-dataset-ex1}.

\begin{table*}
    \begin{center}
  \caption{Comparison of action types in action spotting datasets. *means that the background class (a category that does not belong to the main classes of interest) is not counted as a separate class.}
  \label{tab:comparison-classes}
  \begin{tabular}{llccccc}
    \toprule
    \multicolumn{2}{c}{\textbf{Action}} & \textbf{SoccerNet} & \textbf{SoccerNet-v2} & \textbf{Compr
    Soccer} & \textbf{SSET} & \textbf{SoccerDB} \\
    \midrule
     
     \multicolumn{2}{l}{overhead kick} & \ding{55} & \ding{55} & \ding{51} & \ding{51}  &\ding{55} \\
     
     \multicolumn{2}{l}{solo drive} & \ding{55} & \ding{55} &\ding{51} & \ding{51} &\ding{55}\\
     
     \multicolumn{2}{l}{goal} & \ding{51} & \ding{51} &\ding{51} & \ding{51}& \ding{51}\\
    
     \multirow{ 2}{*}{shot} & on target  & \multirow{ 2}{*}{\ding{55}} & \ding{51} & \multirow{ 2}{*}{\ding{51}} & \multirow{ 2}{*}{\ding{51}} & \multirow{ 2}{*}{\ding{51}}\\

     & off target  &  & \ding{51} & &  & \\
     
     \multicolumn{2}{l}{corner} & \ding{55} & \ding{51} &\ding{51} & \ding{51} & \ding{51}\\
     
     \multirow{ 2}{*}{free kick} & direct & \multirow{ 2}{*}{\ding{55}} & \ding{51} & \multirow{ 2}{*}{\ding{51}} & \multirow{ 2}{*}{\ding{51}} & \multirow{ 2}{*}{\ding{51}} \\
     
     & indirect &  & \ding{51} &  &  & \\
     
     \multicolumn{2}{l}{penalty kick} & \ding{55} & \ding{55} & \ding{51} &\ding{51} &\ding{51}\\
     
     \multirow{ 2}{*}{card} & red & \multirow{ 2}{*}{\ding{51}}&\ding{51} &\ding{51} & \ding{51} & \multirow{ 2}{*}{\ding{51}}\\
     
     & yellow &  & \ding{51} &\ding{51} & \ding{51} &\\
     
     \multicolumn{2}{l}{foul} & \ding{55} &\ding{51} &\ding{51} & \ding{51} & \ding{51}\\
     
     \multicolumn{2}{l}{offside} & \ding{55} & \ding{51} &\ding{51} & \ding{51} & \ding{55}\\
     
     \multicolumn{2}{l}{substitution} & \ding{51} & \ding{51} & \ding{55} & \ding{55}& \ding{51}\\
     
     \multicolumn{2}{l}{ball out of play}  & \ding{55} & \ding{51} & \ding{55} & \ding{55} & \ding{55}\\
     
     \multicolumn{2}{l}{throw-in}  & \ding{55} & \ding{51} & \ding{55} & \ding{55} & \ding{55}\\
     
     \multicolumn{2}{l}{clearance}  & \ding{55} & \ding{51} & \ding{55} & \ding{55} & \ding{55}\\
     
     \multicolumn{2}{l}{kick off}  & \ding{55} & \ding{51} & \ding{55} & \ding{55} & \ding{55}\\
     
     \multicolumn{2}{l}{penalty}  & \ding{55} & \ding{51} & \ding{55} & \ding{55} & \ding{55}\\
     
     \multicolumn{2}{l}{yellow->red card}  & \ding{55} & \ding{51} & \ding{55} & \ding{55} & \ding{55}\\
     
     \multicolumn{2}{l}{injured} & \ding{55}& \ding{55}& \ding{55}&\ding{55} & \ding{51}\\

    \multicolumn{2}{l}{saves} & \ding{55}& \ding{55}& \ding{55}& \ding{55}& \ding{51}\\

    \multicolumn{2}{l}{\textit{corner\&goal}} & \ding{55} & \ding{55} &\ding{51} &\ding{51} &\ding{55}\\
     
     \multicolumn{2}{l}{\textit{corner\&shot}} & \ding{55} & \ding{55} &\ding{51} & \ding{51}&\ding{55}\\
     
     \multicolumn{2}{l}{\textit{free kick\&goal}} & \ding{55}& \ding{55} &\ding{51} & \ding{51} & \ding{55}\\
     
     \multicolumn{2}{l}{\textit{free kick\&shot}} & \ding{55} & \ding{55} &\ding{51} & \ding{51} &\ding{55}\\
     
     \midrule
     \multicolumn{2}{l}{\textbf{\#classes}} & 3 & 17 & 11(+4) & 11(+4) & 10* \\
     \multicolumn{2}{l}{\textbf{Duration [hours]}} & 764 & 764  & 170 & 282 & 669\\
     \multicolumn{2}{l}{\textbf{\#events}} & 6,637 & 110,458 & 6,850 & 10,619 & 37,715 \\
     \multicolumn{2}{l}{\textbf{Freq [\#events/hour]}} & 8.7 & 144.6 & 40.3 & 30.3 & 56.4\\
  \bottomrule
\end{tabular}
  \end{center}
\end{table*}

\begin{figure*}[h]
    \centering
    \includegraphics[width=\textwidth]{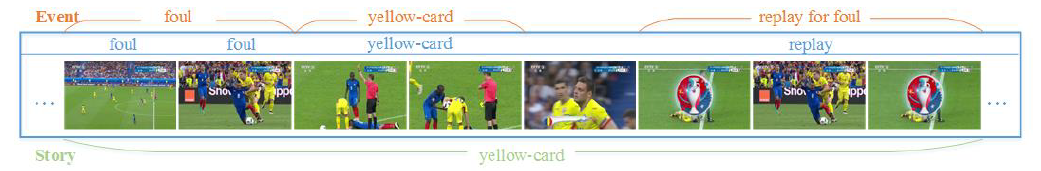}
    \caption{Difference between event and story from Comprehensive Soccer dataset. Source: \citep{comprehensive-soccer-junqing}.}
    \label{fig:comprehensive-soccer-dataset-ex1}
\end{figure*}

\paragraph{Comprehensive Soccer} Comprehensive Soccer dataset\footnote{\url{http://media.hust.edu.cn/dataset/Datasets.htm}}~\citep{comprehensive-soccer-junqing, 10.1007/978-3-030-05716-9_31} is a dataset containing $222$ broadcast soccer videos ($170$ hours in total) in HD 720p ($40\%$ of observations) and 360p ($60\%$ of observations), 25 fps. They notice that most datasets focus on a single task, while a multi-task approach is necessary to analyse sports videos. Their dataset covers three tasks: shot boundary detection (far-view, medium-view, close-view, out-of-field view, playback shot), event detection and player tracking. They divided event annotation into two levels of granularity: event and story proposing $11$ action classes: overhead kick, solo drive, goal, shot, corner, free kick, penalty kick, red card, yellow card, foul, offside, 
 and extra four-story labels: corner\&goal, corner\&shot, free kick\&goal, free kick\&shot. 
 While action describes a single activity, the story provides a comprehensive narrative with contextual background (see Figure~\ref{fig:comprehensive-soccer-dataset-ex1} for more details). 
 They suggest that shot analysis can be essential to action analysis because various views can show different perspectives. For instance, a close-view shot can capture players, coaches and audiences when an event is happening, while far-view present tactics and the arrangement of players in the attacking and defensive team. 


\paragraph{SSET} SSET dataset\footnote{\url{http://media.hust.edu.cn/dataset.htm}}~\citep{sset-na-feng} is an extension of \textbf{Comprehensive Soccer dataset~\citep{comprehensive-soccer-junqing}}. The authors have enriched the previous dataset with 128 recordings and increased the number of event annotations. Finally, the introduced dataset consists of $350$ videos lasting $282$h in total.

\paragraph{SoccerDB} SoccerDB\footnote{\url{https://github.com/newsdata/SoccerDB}}~\citep{soccerdb-yudong} is a dataset with annotations of four tasks: object detection, action recognition, action spotting, and video highlight detection for $343$ soccer matches divided into $171,191$ video segments (some of them are from the SoccerNet dataset). It is worth mentioning that bounding boxes of players and the ball are available but not assigned to player numbers, so this dataset cannot be used for player tracking.

Although SoccerNet, SoccerNet-v2, Comprehensive Soccer, SSET, and SoccerDB are designed for the same task, the defined action labels differ. A comparison of available classes and their statistics can be found in Table~\ref{tab:comparison-classes}.

\paragraph{Soccer-logs} Soccer-logs\footnote{\url{https://
sobigdata-soccerchallenge.it/}} \footnote{\url{https://figshare.com/collections/Soccer\_match\_event\_dataset/4415000}}~\citep{soccer-logs-data-pappalardo} is a large-scale dataset of temporal and spacial soccer events provided by Wyscout. Although they released a huge dataset with over $3$ million events ($100$ times more than the largest open-source dataset SoccerDB), video analysis is hindered because the video files have not been available. Besides events data, authors provide files including annotations of competitions, matches, teams, players, referees and coaches.

\paragraph{SEV dataset} It~\footnote{\url{https://github.com/FootballAnalysis/footballanalysis}}~\citep{karimi2021soccer} consists of $42000$ event-related images split into train, test, and validation data. The dataset includes annotations for 7 soccer events: corner kick, penalty kick, free kick, red card, yellow card, tackle, and substitute.

\paragraph{EIGD-S} EIGD-S~\footnote{\url{https://github.com/mm4spa/eigd}}~\citep{10.1145/3475722.3482792} is a dataset consisting of five soccer matches recordings with gold standard annotations for $125$ minutes. The dataset includes multiple annotations for two matches from 4 experts and one inexperienced annotator. URLs of videos link to YouTube, where videos with audio paths are available. Annotation was prepared according to the proposed taxonomy assuming the hierarchical structure of events~\citep{10.1145/3475722.3482792}. Unlike the other datasets, EIGD-S contains annotations of high-level events, such as passes along with low-level events, including goals or cards.

\paragraph{VisAudSoccer} \citet{9106051} proposed a new dataset (here denoted as \textbf{VisAudSoccer}), which contained data from $460$ soccer game broadcasts, including $300$ videos downloaded from SoccerNet~\citep{soccernet}, lasting about $700$ hours in total in video format. Audio data is available for $160$ games with commentator voices categorized as "excited" and "not-excited". Events are divided into four classes: celebrate ($1320$ events), goal/shoot ($1885$ events), card ($2355$ events), and pass ($2036$ events). The dataset is not publicly available.

\paragraph{SoccerSummarization}\citet{Gautam2022Oct}~\footnote{\url{https://github.com/simula/soccer-summarization}} extended SoccerNet-v2~\citep{soccernet-v2} with news, commentaries and lineups from BBC.

\subsection{Multi-Sports Datasets}


    
    


    
\begin{figure}
    \includegraphics[width=.2\textwidth]{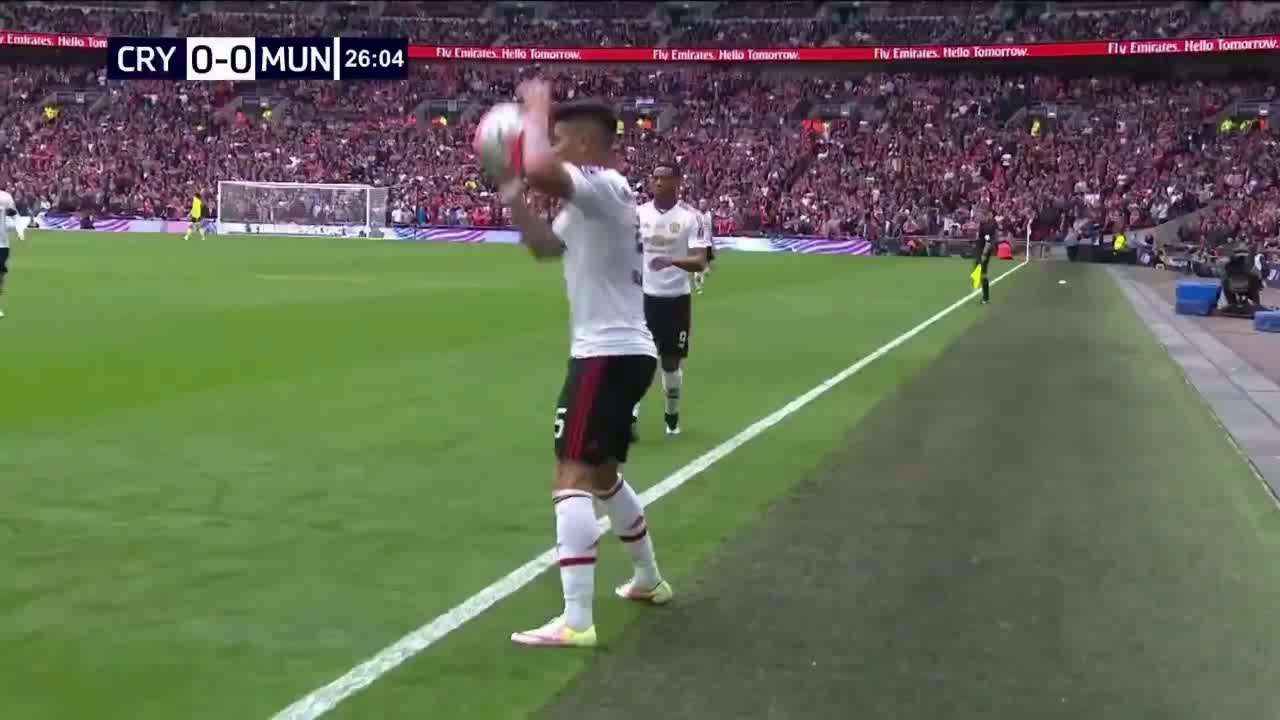}\hfill
    \includegraphics[width=.2\textwidth]{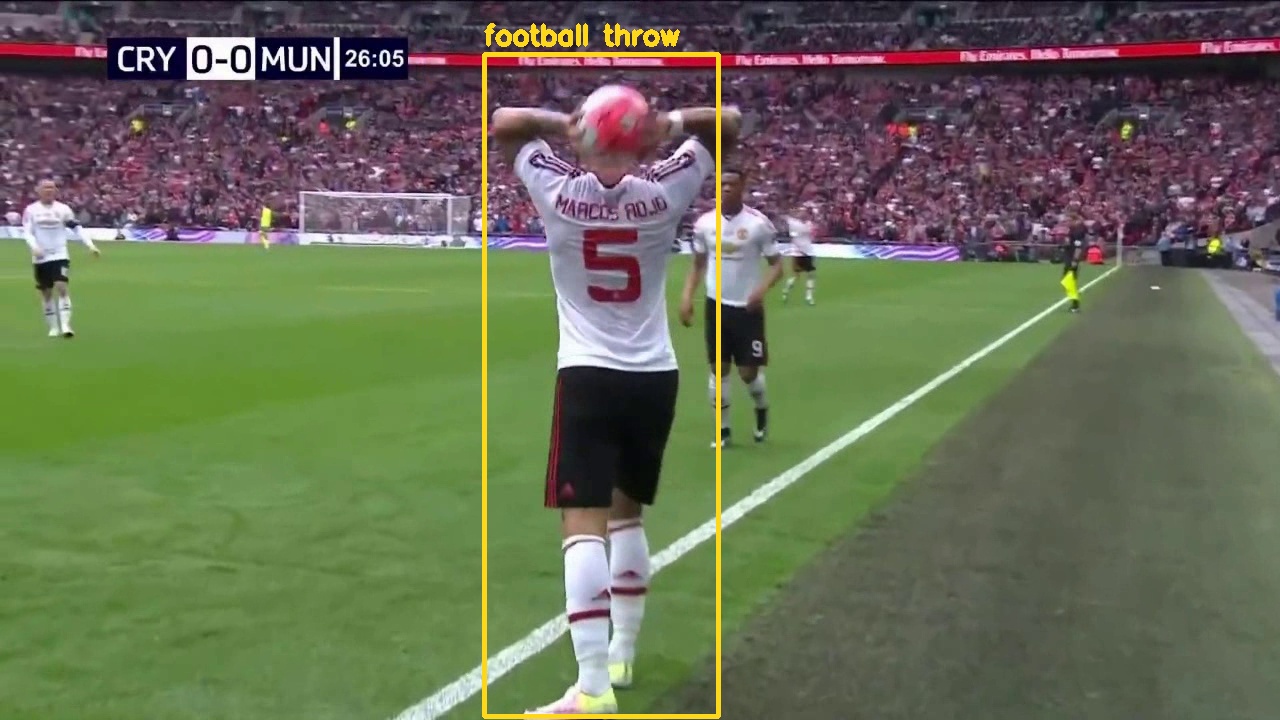}\hfill
    \includegraphics[width=.2\textwidth]{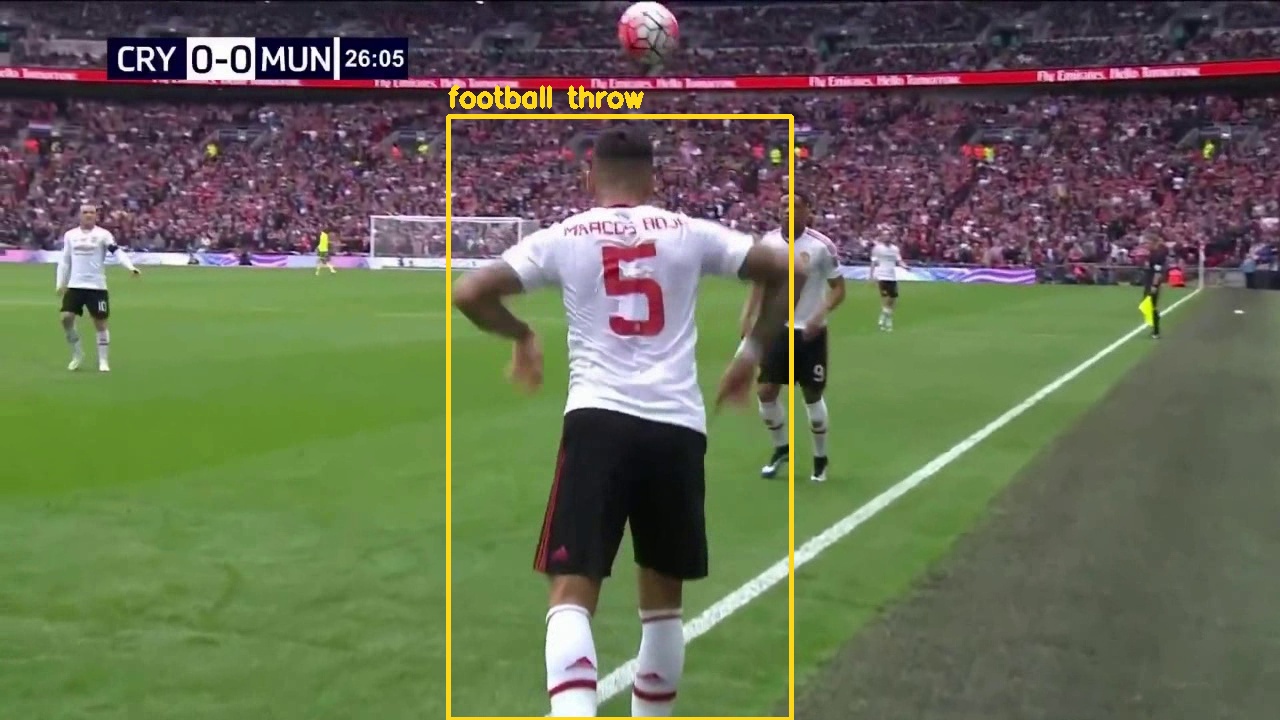}\hfill
    \includegraphics[width=.2\textwidth]{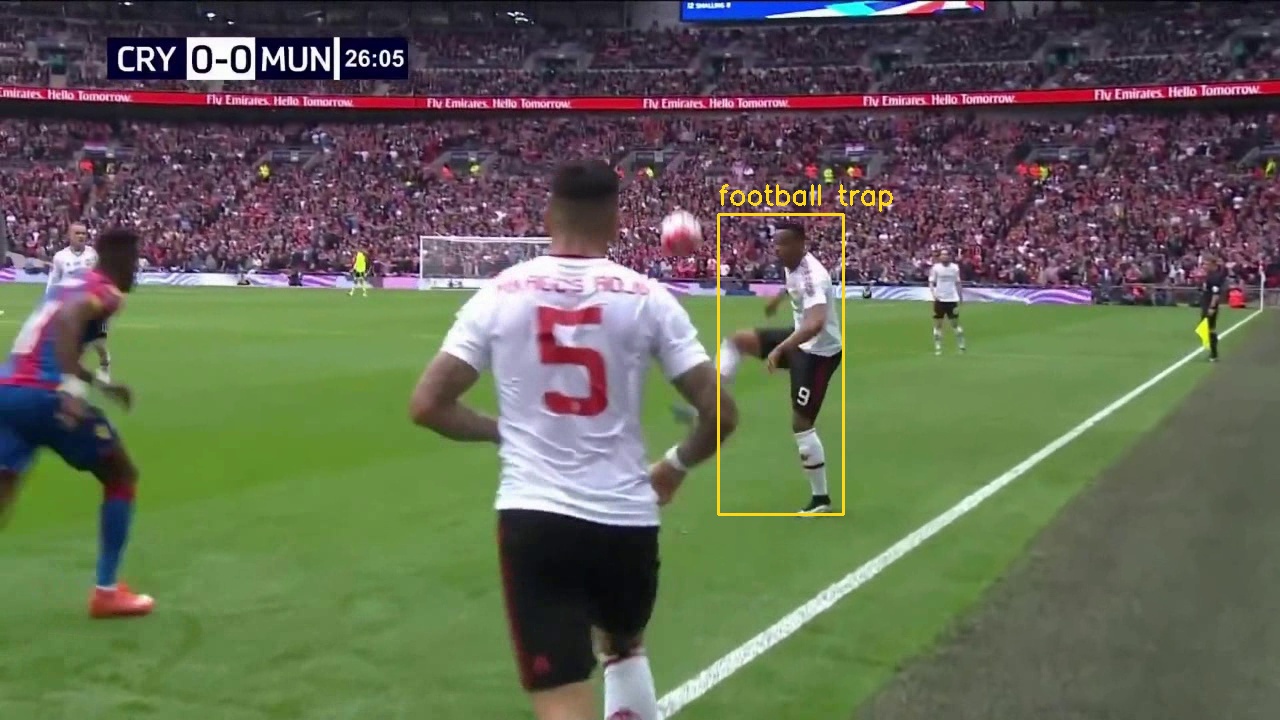}\hfill
    \includegraphics[width=.2\textwidth]{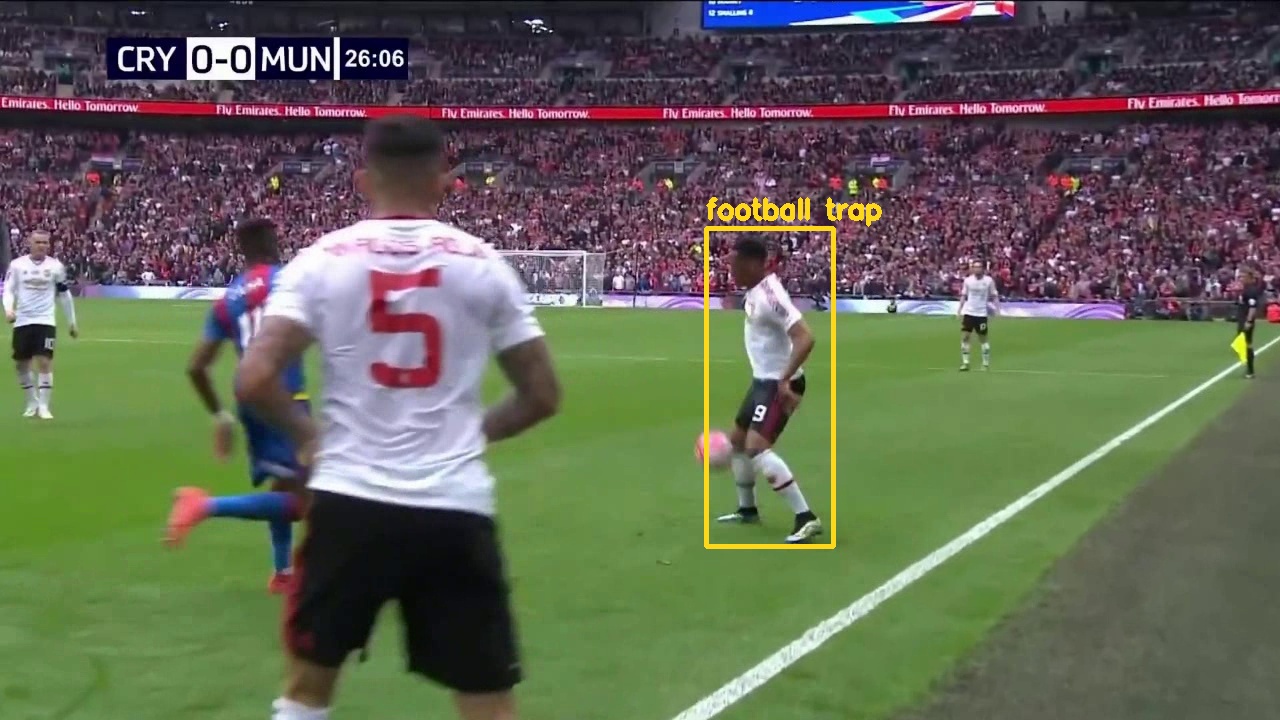}
    \includegraphics[width=.2\textwidth]{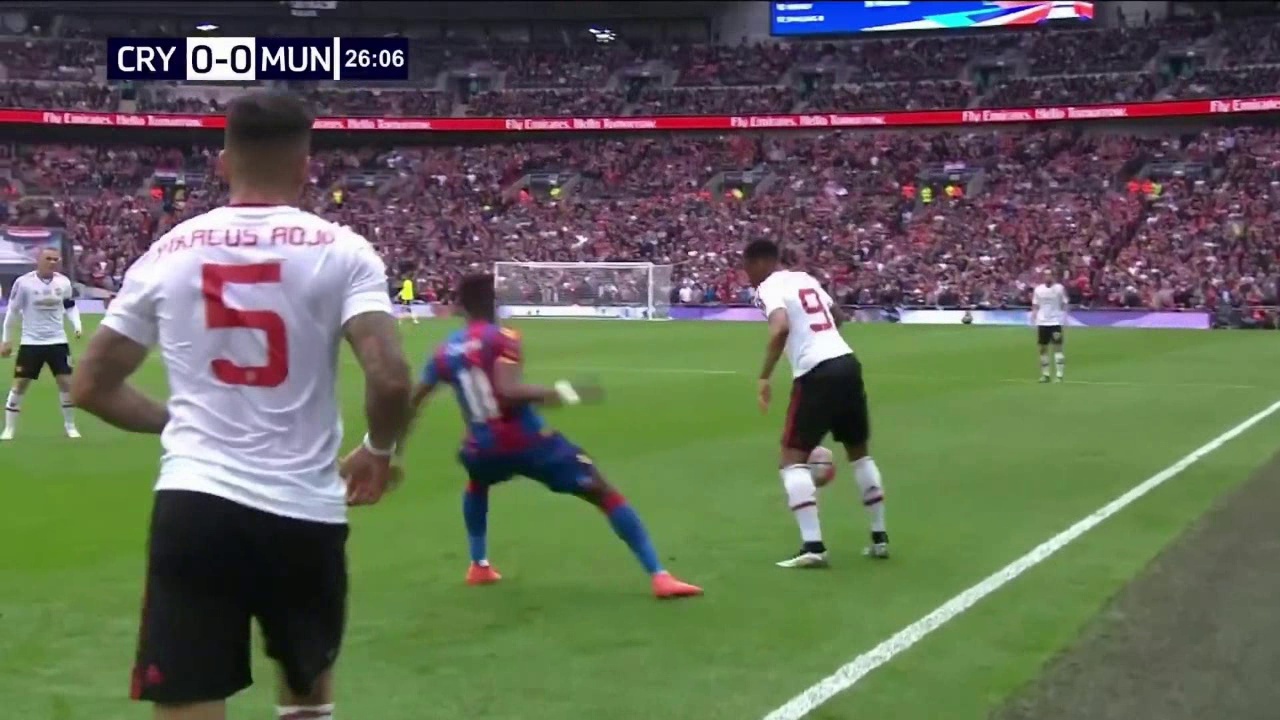}\hfill
    \includegraphics[width=.2\textwidth]{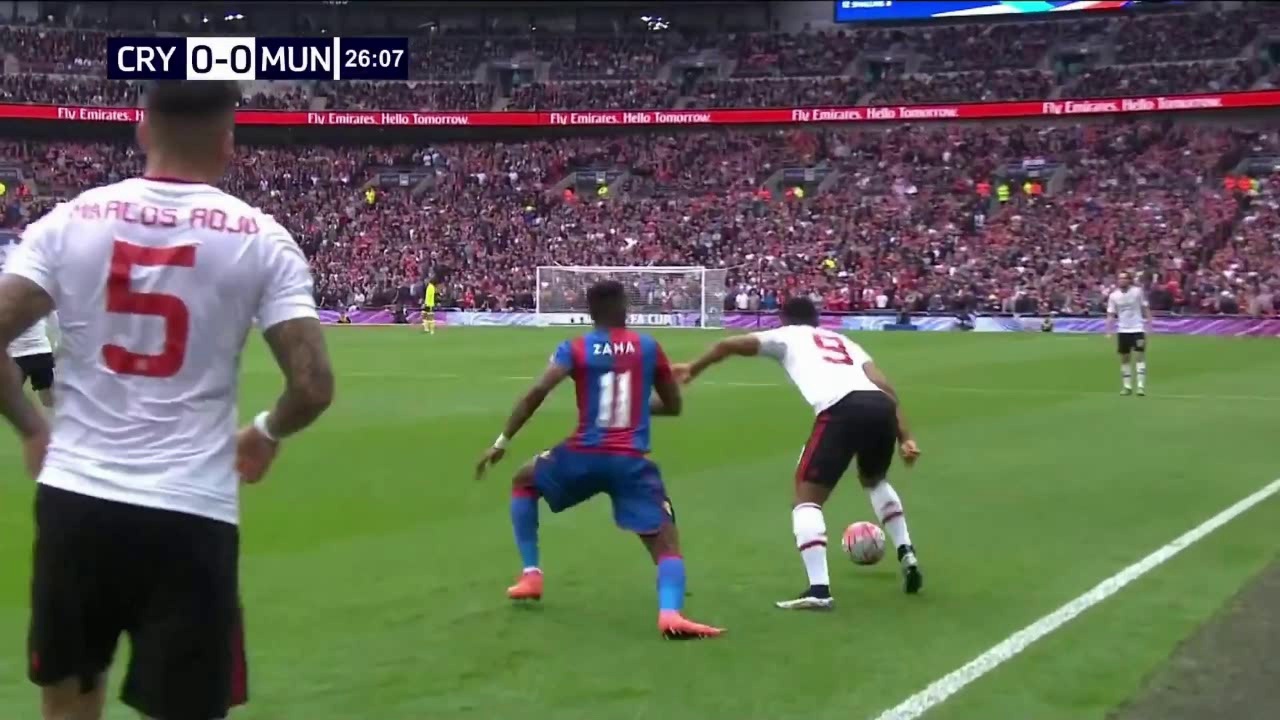}\hfill
    \includegraphics[width=.2\textwidth]{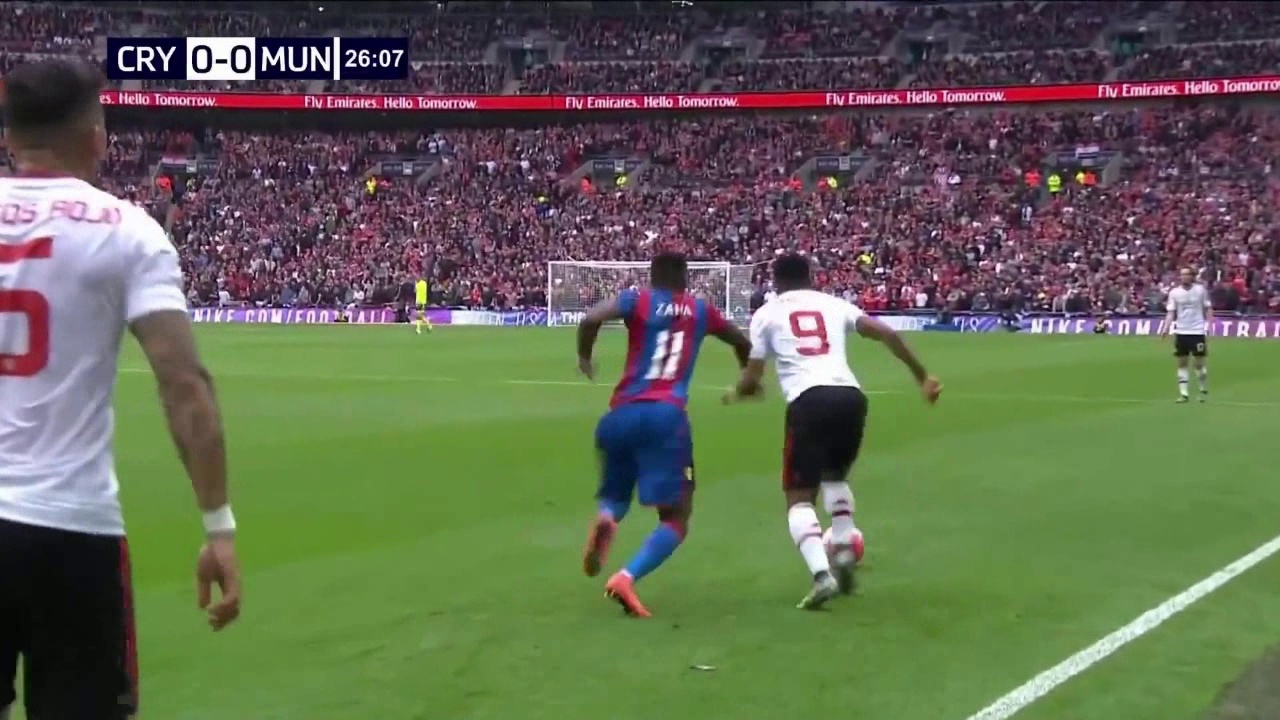}\hfill
    \includegraphics[width=.2\textwidth]{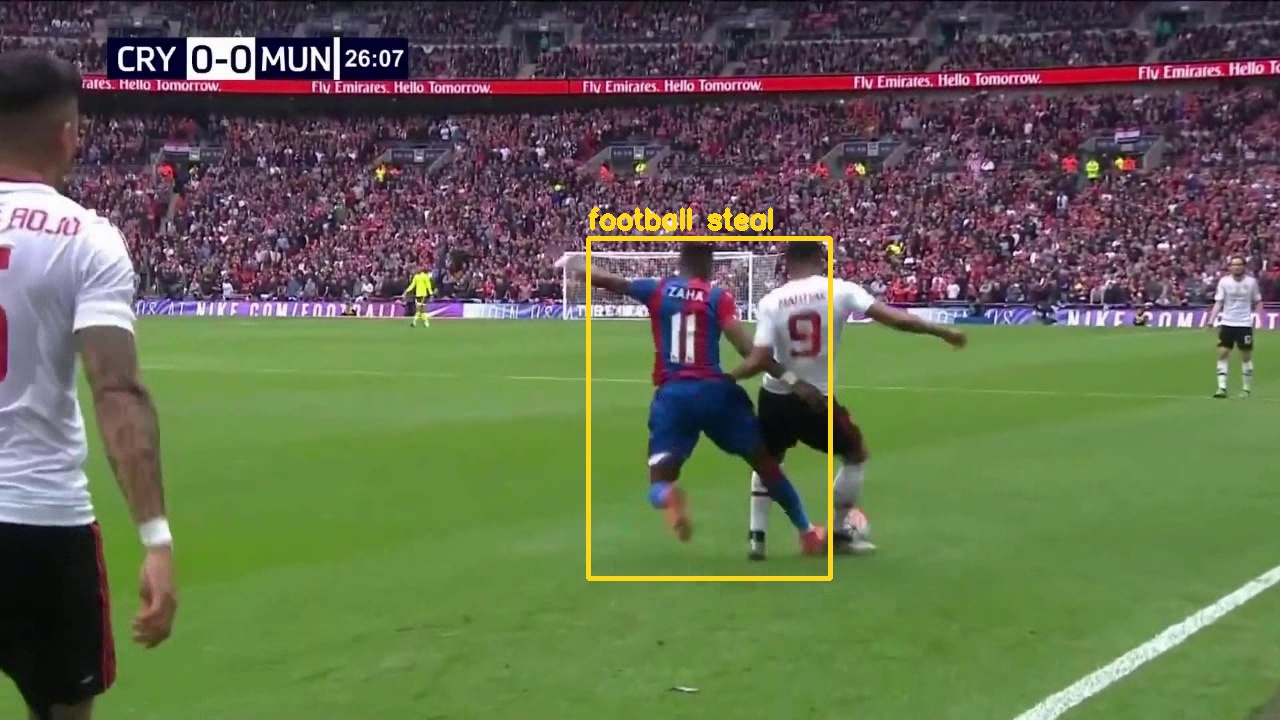}\hfill
    \includegraphics[width=.2\textwidth]{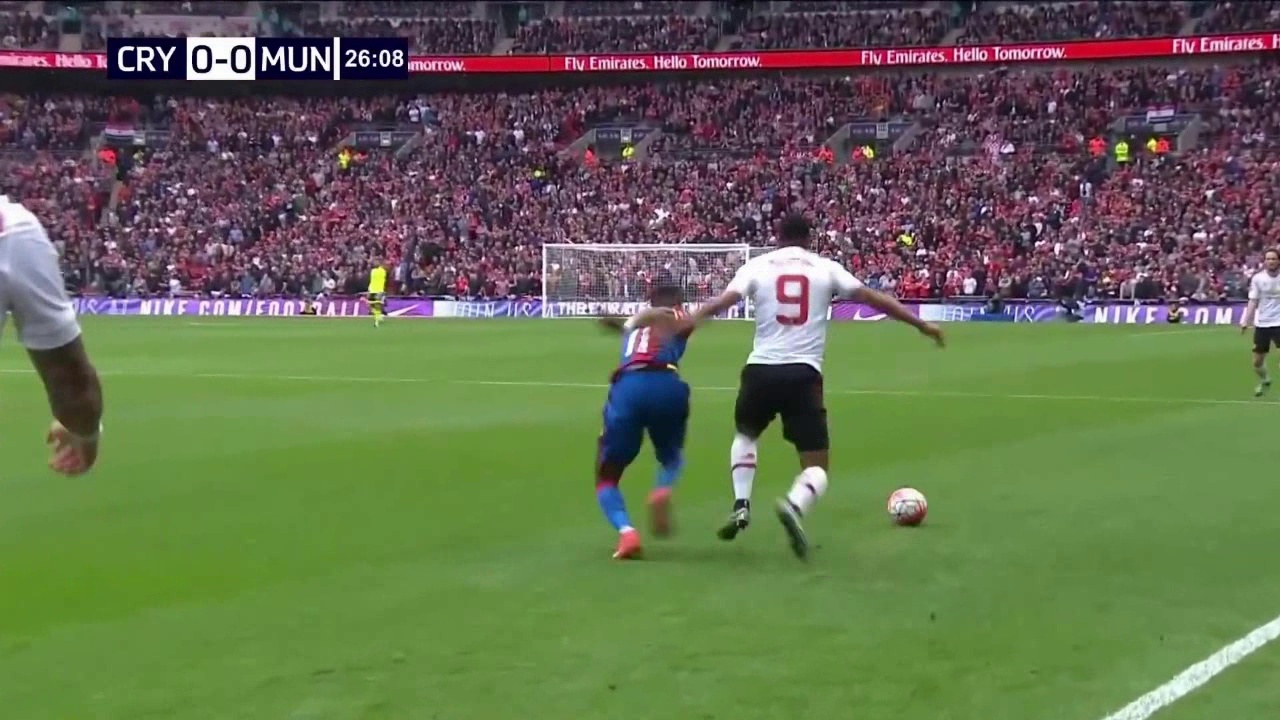}\hfill
    
    \caption{Examples of annotations from MultiSports~\citep{multisports-li-yixuan} dataset. Players participating in a given action are annotated with bounding boxes.}\label{fig:multisports_dataset_ex}
\end{figure}

    
\paragraph{MultiSports} MultiSports\footnote{\url{https://deeperaction.github.io/datasets/multisports.html}}~\citep{multisports-li-yixuan} released spatio-temporal multi-person action detection dataset for basketball, volleyball, soccer, and aerobic gymnastics. After consulting with athletes, they proposed $66$ action labels, e.g. soccer pass, trap, defence, tackle, and long ball. Additionally, a handbook was created to define actions and their temporal boundaries. Videos were downloaded from YouTube and then trimmed into shorter clips. In the end, $800$ clips are available for each sport, which amounts to around $5$ hours of recordings for soccer. The number of relevant action categories for soccer equals $15$, and $12,254$ instances were annotated. The authors emphasize that their dataset  differs from other datasets due to its complexity, high quality, and diversity of videos. In the case of soccer, it is the first publically available dataset that contains spatio-temporal action annotations. Examples of MultiSports annotations can be found in Figure~\ref{fig:multisports_dataset_ex}.

\subsection{Other Datasets}

\paragraph{UCF Sports} UCF Sports\footnote{\url{https://www.crcv.ucf.edu/data/UCF\_Sports\_Action.php}}~\citep{ucf-sports-0-dataset-soomro-khurram, ucf-sports-dataset-soomro-khurram} is a dataset for spatio-temporal action recognition with $10$ possible classes: diving, golf swing, kicking, lifting, riding a horse, running, skateboarding, swinging-bench, swinging-side, and walking. $150$ videos lasting about $6.39$ seconds were gathered from broadcast television networks like ESPN and BBC. UCF Sports is known as one of the first datasets that published not only action class annotations but also bounding boxes of areas associated with the action. It was broadly used to conduct action classification experiments~\citep{TangentBundle2011, DTM2010, AFMKL011, ScalableActionRecognitionSubSpaceForest2012} and spatio-temporal action recognition. It differs a lot from the task described in MultiSports dataset~\citep{multisports-li-yixuan} where the video is not temporally trimmed, multiple players and actions can be detected, and single action occurs only in a small subset of time. However, UTF-Sports initiated the advancement of this domain and inspired authors to develop interesting solutions. Due to the fact that this dataset contains only a few events related to soccer (kicking and running), the results and methods have not been widely described in this article.


\paragraph{UCF-101} The UCF-101~\citep{Soomro2012UCF101AD} dataset was introduced as an action recognition dataset of realistic action videos from YouTube. $2$ out of $101$ action categories are related to soccer: soccer juggling and soccer penalty. Other action categories include diving, typing, bowling, applying lipstick, knitting etc.

\paragraph{Goal} \citet{goal-event-sport-2017-tsagkatakis} proposed a dataset to classify a single \textit{goal} class in soccer. The dataset consists of videos from YouTube: $200$ 2-3 second-long videos for \textit{goal} class and $200$ videos for \textit{no-goal} class.

\paragraph{Offside Dataset} A dataset\footnote{\url{https://github.com/Neerajj9/Computer-Vision-based-
Offside-Detection-in-Soccer}}~\citep{offside_dataset}  that can be used to assess the effectiveness of a methodology for offside detection. It consists of about 500 frames that are publicly available. The authors highlight that this dataset has been carefully curated to include a diverse range of soccer match scenes demonstrating the different challenges such a system may encounter.

\begin{figure}
\centering
    \includegraphics[width=.27\textwidth]{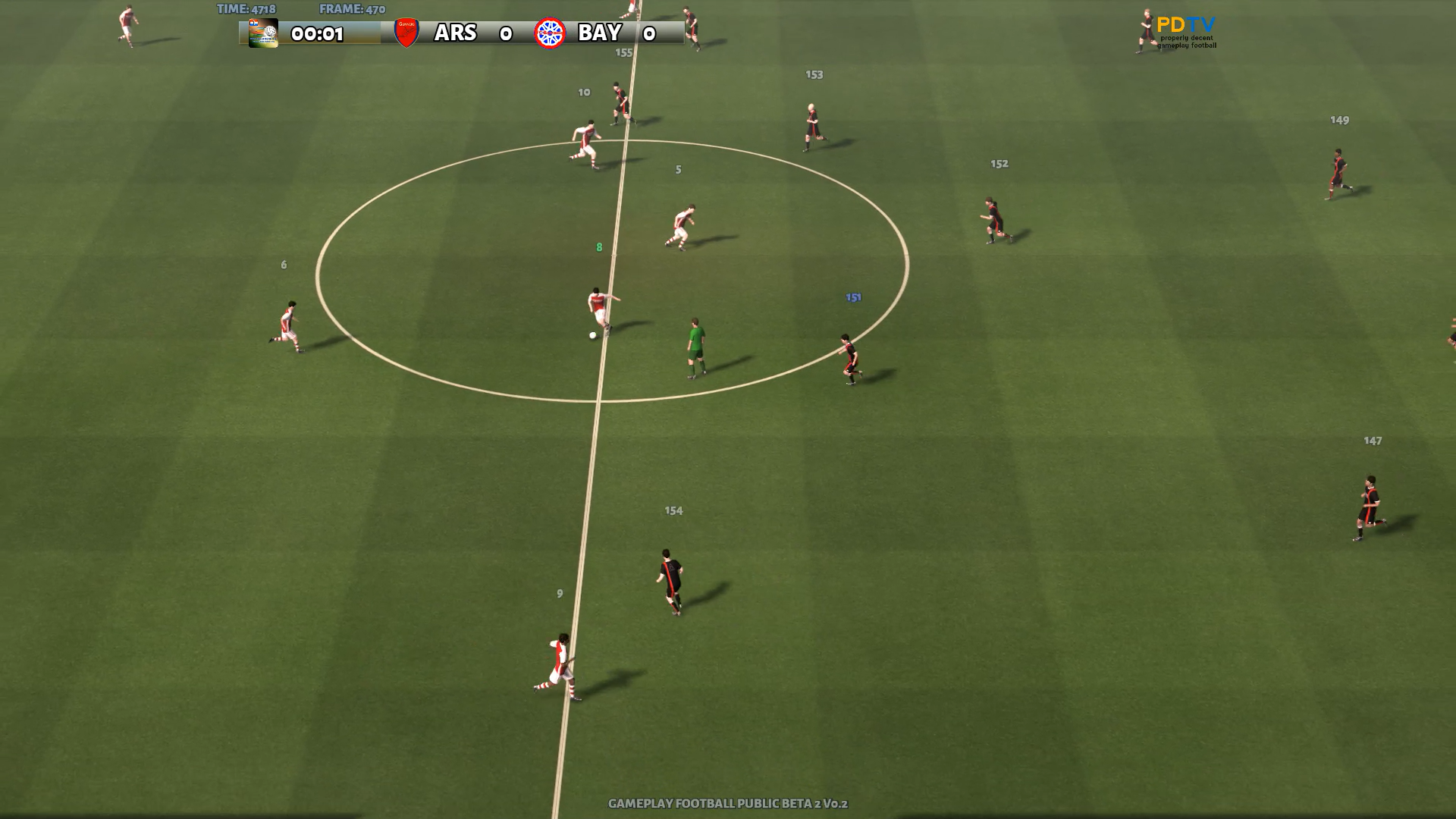}\hfill
    \includegraphics[width=.27\textwidth]{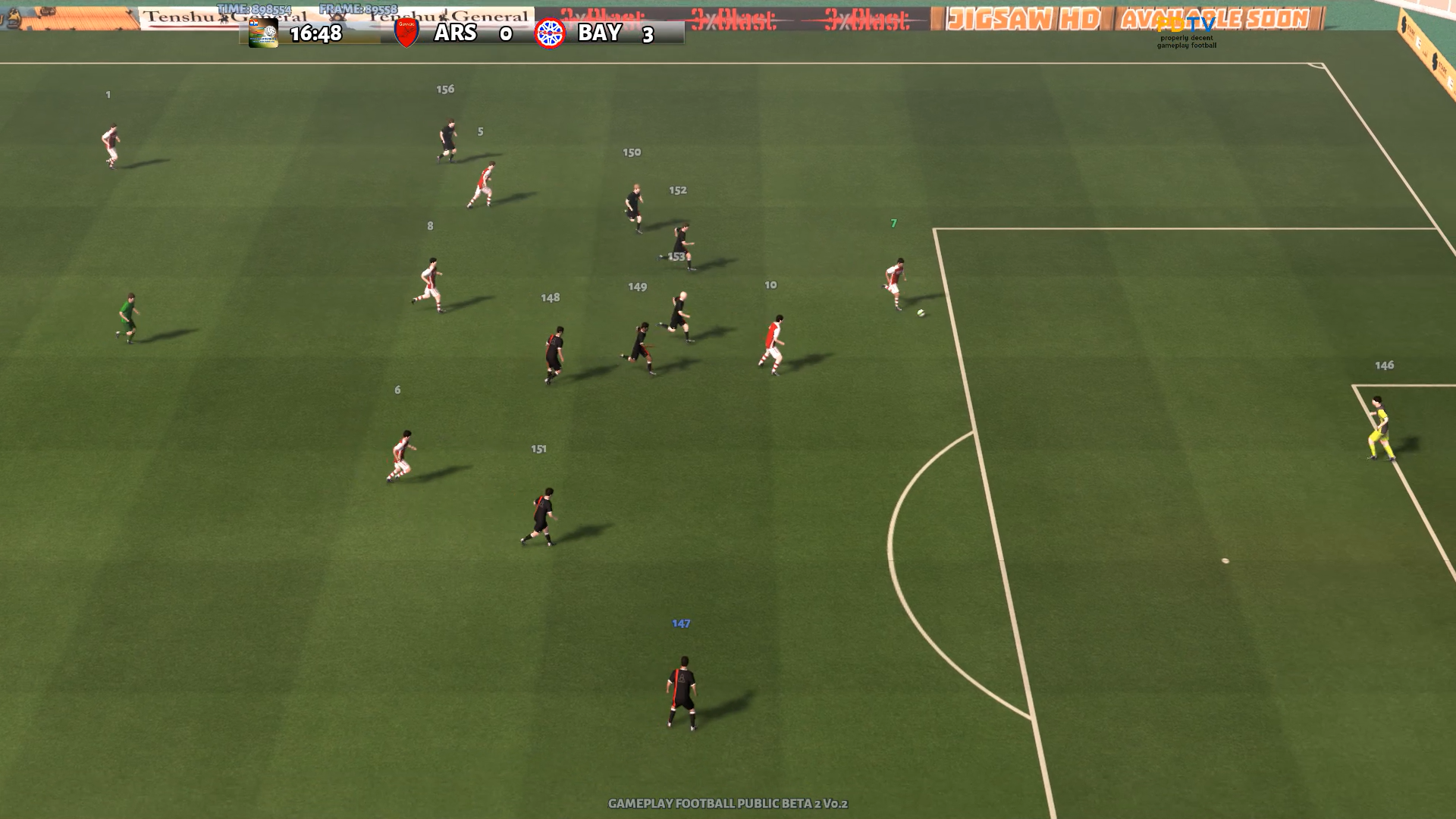}\hfill
    \includegraphics[width=.27\textwidth]{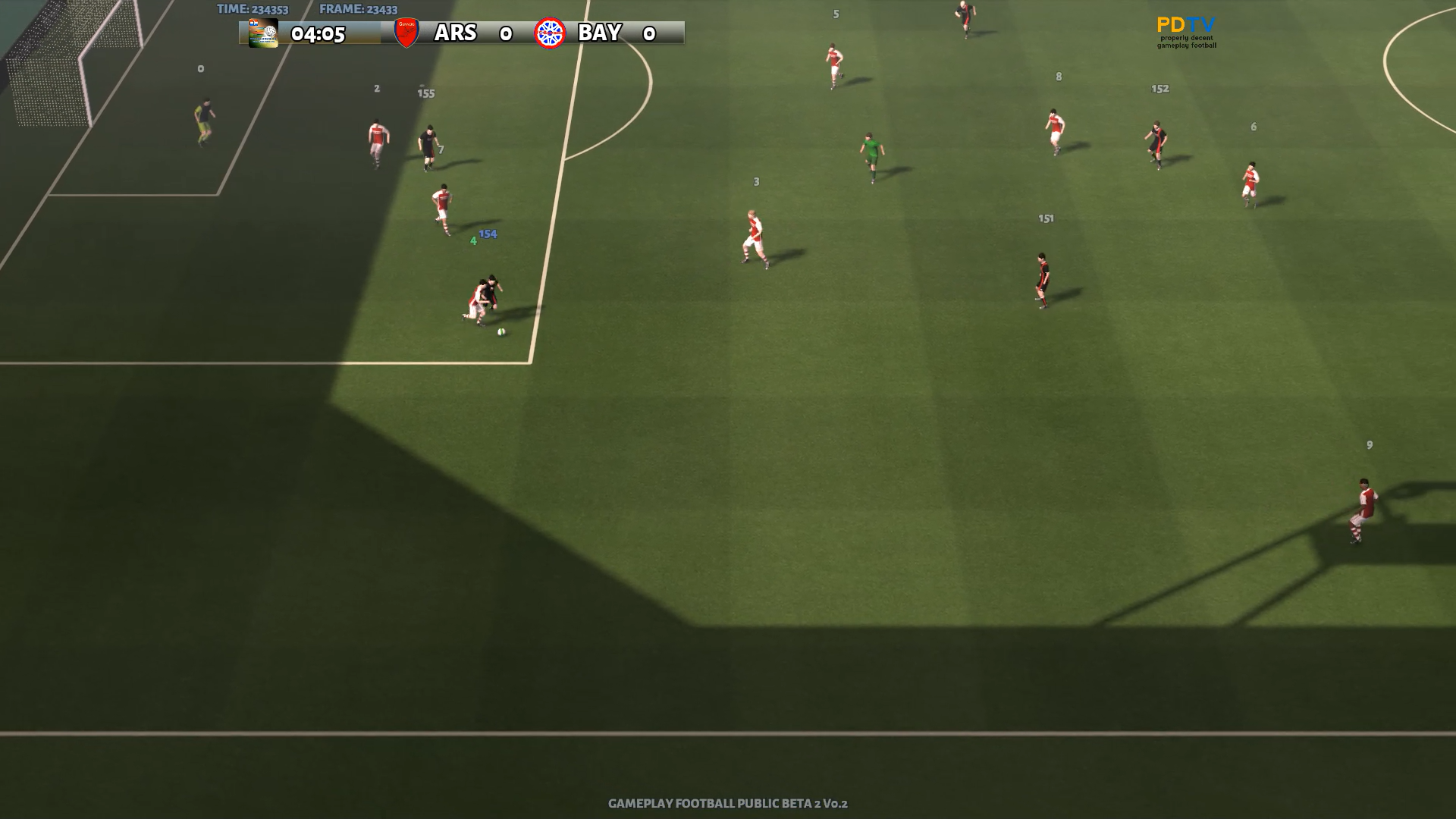}\hfill
    \includegraphics[width=.27\textwidth]{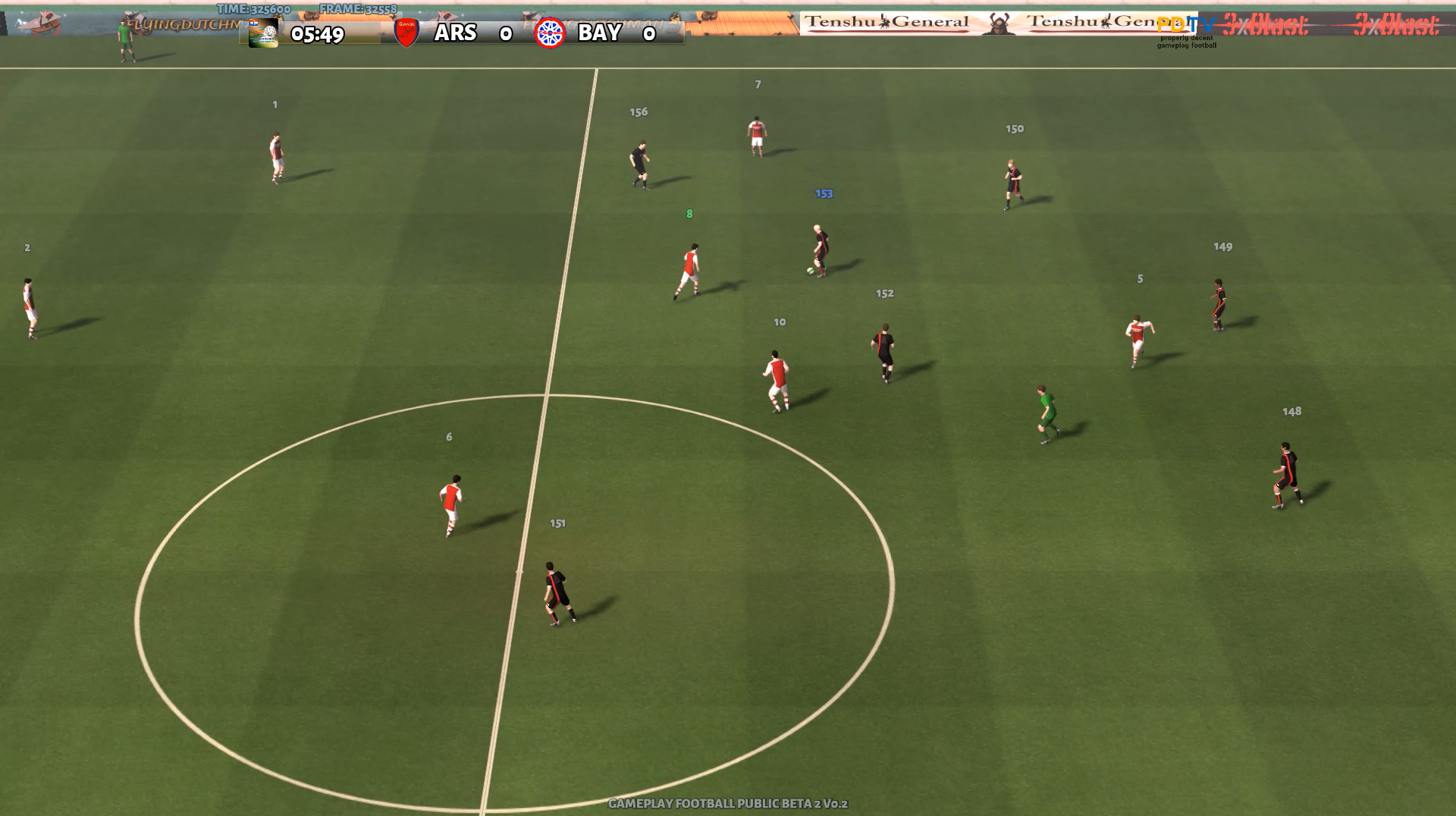} \hfill

    \caption{Examples from artificially generated  SoccER dataset~\citep{Slicing_Dicing_Soccer_2020}. }\label{fig:soccer_dataset_ex}
\end{figure}

\paragraph{SoccER} Soccer Event Recognition~\footnote{\url{https://gitlab.com/grains2/slicing-and-dicing-soccer}}~\citep{Slicing_Dicing_Soccer_2020} is a synthetic dataset consisting of $500$ minutes of game recordings gathered from the open source Gameplay Football engine that can be an approximation of real game. $1.6$ million atomic events and $9,000$ complex events are annotated. Atomic events (kicking the ball, ball possession, tackle, ball deflection, ball out, goal, foul, penalty) are spatio-temporally annotated. Complex events occur over a wide area, involve multiple participants, or can be composed of multiple other events (pass, pass then goal, filtering pass, filter pass then goal, cross, cross then goal, tackle, shot, shot then goal, saved shot). Examples can be found in Figure~\ref{fig:soccer_dataset_ex}. 

\paragraph{GOAL} GrOunded footbAlL commentaries (\textbf{GOAL}), dataset~\citep{Suglia2022GoingFG} contains $1107$ game recordings transcribed to text. Although this dataset is dedicated to tasks such as commentary retrieval or commentary generation, its contents can also be valuable as an additional modality in action recognition.






\section{Methods}
\label{sec:methods}









\subsection{Action Recognition and Spotting}

\begin{table*}
\begin{center}
  \caption{Methods used for action recognition in analysed articles. Features are represented as \tikzcircle[green, fill=green]{3pt} - image, \tikzsquare[red, fill=red]{3pt} - audio
  .}
  \label{tab:methods-analysed-articles-classification}
  \centering
  \begin{tabular}{p{3.4cm}cp{3cm}ccc}
    \toprule
    \textbf{Article} & \textbf{Dataset} & \textbf{Method} & \textbf{Features} & \textbf{mAP} & \textbf{Top-1 Acc}  \\
    \midrule
     \citep{soccernet} & SoccerNet & AvgPool &  \tikzcircle[green, fill=green]{3pt} & 40.7 & -\\
      \citep{soccernet} & SoccerNet & MaxPool &  \tikzcircle[green, fill=green]{3pt}  & 52.4 & -\\
     \citep{soccernet} & SoccerNet & NetVLAD &  \tikzcircle[green, fill=green]{3pt} &67.8 & -\\
     \citep{soccernet} & SoccerNet & NetRVLAD &  \tikzcircle[green, fill=green]{3pt}  & 67.4 & -\\
     \citep{soccernet} & SoccerNet & NetFV &  \tikzcircle[green, fill=green]{3pt}  & 64.4 & -\\
     \citep{soccernet} & SoccerNet & SoftBOW &  \tikzcircle[green, fill=green]{3pt}  & 62.0 & -\\
     \citep{Vanderplaetse2020ImprovedSA} & SoccerNet & AudioVid &  \tikzcircle[green, fill=green]{3pt} \tikzsquare[red, fill=red]{3pt} & 73.7 & -\\
     \citep{multimodal-transformer-Gan-Yaozong-2022} 
     & SoccerNet-v2 & PM & \tikzcircle[green, fill=green]{3pt} \tikzsquare[red, fill=red]{3pt} & - & 62.4\\
    \hline
    \citep{9106051} & 
    VisAudSoccer & I3D~\citep{carreira2017quo} & \tikzcircle[green, fill=green]{3pt}  & 95.2 & 90.1\\
    \citep{9106051} & 
    VisAudSoccer & I3D-NL~\citep{wang2018non} & \tikzcircle[green, fill=green]{3pt}  & 96.9 & 92.5 \\
    \citep{9106051} & 
    VisAudSoccer & ECO~\citep{zolfaghari2018eco} & \tikzcircle[green, fill=green]{3pt}  & 96.3 & 92.2 \\
    \citep{9106051} & 
    VisAudSoccer & SlowFast~\citep{slowfast_Feichtenhofer2018SlowFastNF} & \tikzcircle[green, fill=green]{3pt}  & 95.1 & 88.1 \\
  \bottomrule
\end{tabular}
\end{center}
\end{table*}


\begin{table*}
  \caption{Methods used for action spotting in analysed articles.  Features are represented as \tikzcircle[green, fill=green]{3pt} - image, \tikzsquare[red, fill=red]{3pt} - audio
  , \tikzcircle[orange, fill=orange]{3pt} - graph. * denotes that model was evaluated on the challenge dataset.}
  \label{tab:methods-analysed-articles-spotting}
  \begin{tabular}{p{3cm}cp{2.5cm}ccp{1.5cm}c}
    \toprule
    \textbf{Article} 
    & \textbf{Dataset} & \textbf{Method} & \textbf{Features} & \textbf{Avg mAP} & \textbf{Tight Avg mAP} &  \textbf{Acc} \\
    \midrule

     \citep{soccernet-v2}  & SoccerNet & NetVLAD & \tikzcircle[green, fill=green]{3pt}  & 49.7 & - & -\\
     \citep{Vanderplaetse2020ImprovedSA}  & SoccerNet & AudioVid & \tikzcircle[green, fill=green]{3pt} \tikzsquare[red, fill=red]{3pt} & 56.0 & - & -\\
    \citep{calf-Cioppa2020Context}  & SoccerNet & CALF & \tikzcircle[green, fill=green]{3pt} & 62.5 & - & -\\
    \citep{Vats2020EventDI}  & SoccerNet & MTTCNN & \tikzcircle[green, fill=green]{3pt} &  60.1 & - & -\\
    \citep{realtime-det-3d-rongved-2020}  & SoccerNet &  3dCNN & \tikzcircle[green, fill=green]{3pt} &  32.0 & - & -\\
    \citep{soccer-self-attention-2020}  & SoccerNet & NetVLAD + self-attention  & \tikzcircle[green, fill=green]{3pt} & - & - & 74.3\\
    \citep{Tomei2021RMSNetRA}  & SoccerNet & RMS-Net & \tikzcircle[green, fill=green]{3pt} &  65.5 & - & -\\
    \citep{NergrdRongved2021AutomatedED} & SoccerNet & 2D-CNN AudVid  & \tikzcircle[green, fill=green]{3pt} \tikzsquare[red, fill=red]{3pt} & - & - &90.85 \\
    \citep{Mahaseni2021SpottingFE} & SoccerNet &  CNN + Dilated RNN & \tikzcircle[green, fill=green]{3pt}  & 63.3 & - & -\\
    \citep{Multiple-Scene-Encoders-Shi-Yuzhi-2022} & SoccerNet & Multiple Scene Encoder & \tikzcircle[green, fill=green]{3pt} & 66.8 & - & -\\
    \citep{9959985}  & SoccerNet & CNN-GRU metric learning & \tikzcircle[green, fill=green]{3pt} &  64.9 & - & -\\
    
    \hline
    
    \citep{soccernet-v2}  & SoccerNet-v2 & MaxPool & \tikzcircle[green, fill=green]{3pt} &  18.6& - & -\\
    \citep{soccernet-v2}  & SoccerNet-v2 & NetVLAD & \tikzcircle[green, fill=green]{3pt} &  31.4& - & -\\
    \citep{soccernet-v2}  & SoccerNet-v2 & AudioVid & \tikzcircle[green, fill=green]{3pt} \tikzsquare[red, fill=red]{3pt} &  40.7& - & -\\
    \citep{soccernet-v2}  & SoccerNet-v2 & CALF & \tikzcircle[green, fill=green]{3pt} &  41.6& - & -\\
    \citep{feature_combination_baidu_attention_xin_zhou_2021} & SoccerNet-v2 & Vidpress Sports & \tikzcircle[green, fill=green]{3pt} &  74.1 & - & - \\
    \citep{Giancola2021TemporallyAwareFP}  & SoccerNet-v2 & NetVLAD++ & \tikzcircle[green, fill=green]{3pt} & 53.4& - & -\\
    \citep{Transformer-Based-action-spotting-2022}  & SoccerNet-v2 & transformer & \tikzcircle[green, fill=green]{3pt} & 52.04* & - & - \\
    \citep{Cioppa2021CameraCA}  & SoccerNet-v2 & CC+RN+FCL & \tikzcircle[green, fill=green]{3pt} \tikztriangle[orange, fill=orange]{3pt} &  46.8& - & - \\
    
    \citep{graph-for-action-spotting-cartas-2022} & SoccerNet-v2 & RGB+Audio+ Graph & \tikzcircle[green, fill=green]{3pt} \tikzsquare[red, fill=red]{3pt} \tikztriangle[orange, fill=orange]{3pt} & 57.8 & - & - \\

    \citep{ste-darwish-abdulrahman-2022} & SoccerNet-v2 & STE & \tikzcircle[green, fill=green]{3pt} & 74.1 & 58.5 & -\\
    \citep{Multiple-Scene-Encoders-Shi-Yuzhi-2022} & SoccerNet-v2 & Multiple Scene Encoder & \tikzcircle[green, fill=green]{3pt} &  75.3& - & -\\
    \citep{spotformer-cao-menggi-2022}  & SoccerNet-v2& SpotFormer & \tikzcircle[green, fill=green]{3pt} &  76.1 &60.9 & -\\

    \citep{Hong2022SpottingTP} & SoccerNet-v2 & E2E-Spot 800MF & \tikzcircle[green, fill=green]{3pt} & 74.1 &  61.8 & -\\
    \citep{faster-tad-2022} & SoccerNet-v2 & Faster-TAD & \tikzcircle[green, fill=green]{3pt}  & - & 54.1 & -\\
    
    \citep{Soares2022TemporallyPA} & SoccerNet-v2 & DU+SAM+mixup & \tikzcircle[green, fill=green]{3pt} & 77.3 &60.7 & -\\
    \citep{AS_Dense_Detection_Anchors_RevisitedSoares2022} & SoccerNet-v2 & DU+SAM+mixup +Soft-NMS & \tikzcircle[green, fill=green]{3pt} & 78.5 &65.1 & -\\

  \bottomrule
\end{tabular}
\end{table*}

Action analysis in soccer has been an important task and has attracted many researchers. The first articles extracted video features and, based on that, classified clips into predefined categories using rule-based algorithms or classical machine learning models. 

\citet{Soccer_Event_Detection_2018} experimented with a short 5-minute long video, where events (ball possession and kicking) were classified with a rule-based system. The event detector took as an input bounding boxes of ball  with associated confidence scores. Similarly, a rule-based system consulted with soccer experts was proposed in~\citep{Recognizing_Events_in_Spatiotemporal_Soccer_Data_2020_khaustov} to classify events such as ball possession, successful and unsuccessful passes, and shots on goal. It was evaluated on two datasets from Data Stadium and Stats Perform.

Initially, models relied mainly on feature engineering extracting semantic concepts in clips~\citep{Exciting_Event_Detection_Incremental_Learning_2005, fuzzy_HOSSEINI2013846, bayesian_network_2015, Raventos2015-vt, Bayesian_Network_Copula_2014, xie2011novel}. Colour, texture and motion are represented. Also, representation is enriched with mid-level features, including camera view labels, camera motion, shot boundary descriptions, object detections, counting players, grass ratio, play-break segmentation, dominant colour, or penalty area. Audio descriptors such as whistle information or MPEG audio features 
are also used~\citep{fuzzy_HOSSEINI2013846, bayesian_network_2015, Raventos2015-vt, kapela2015real, 1199134, 1221333}. Particularly, audio keywords such as long-whistling, double-whistling (indicating foul), multi-whistling, excited commentator speech, and excited audience sounds can assist in detecting events such as a free kick, penalty kick, foul, and goal in soccer~\citep{1221608}.
These features are fed to classifiers, such as SVM~\citep{Exciting_Event_Detection_Incremental_Learning_2005, 7486522, 1512240}, Hidden Markov models (HMM)~\citep{HMM_2011_qian, 6727163, Sports_highlight_detection_keyword_sequences_HMM_2004, Xiong2005-ar, 5709225, Semantic_Indexing_2004, 10.1007/978-3-642-15696-0_41}, bayesian networks~\citep{Bayesian_Network_Copula_2014,1658037}, hierarchical Conditional Random Field~\citep{hierarchical_crf_2009}, or fuzzy logic~\citep{8015426}.
Along with the development of science and access to better computing machines, video representation improved (VGG-16 backbone~\citep{10.1007/978-3-030-05716-9_31}), and classifiers became more complex, e.g. Long-short Term Memory (LSTM)~\citep{Fakhar2019-zy, football-actions-tsunoda, combined-rnn-cnn-2016-jiang, 10.1007/978-3-030-05716-9_31}, CNN~\citep{end-to-end-soccer-2018-hong, Khan2018LearningDC, combined-rnn-cnn-2016-jiang}, or GRU~\citep{combined-rnn-cnn-2016-jiang}.

A very interesting approach was investigated in~\citep{4668533, sport-tag-2011}, where data published on the Internet, including Twitter posts, were used to identify events in various games (like soccer and rugby). In~\citep{8621906}, the authors use the live text of soccer matches as additional input to the model. Text model composed of TextCNN~\citep{kim-2014-convolutional}, LSTM with attention~\citep{yang-etal-2016-hierarchical} and VDCNN~\citep{conneau-etal-2017-deep} detect events in time and classify them. Then, Optical Character Recognition (OCR) links video time to associated texts. If necessary, a video-model is employed to detect events. Another noteworthy method was proposed by~\citet{events-tracing-data-2022}, who utilised tracking data and a tree-based algorithm to detect events. A similar solution was suggested in~\citep{richly2016recognizing} where positional data was employed to feed event classifiers such as SVM, K-Nearest Neighbors and Random Forest. 



In \citep{soccernet}, authors introduced SoccerNet dataset together with benchmarks for action classification and spotting tasks. They achieved an average-mAP of 49.7\% for a threshold ranging from 5 to 60 seconds in the spotting task. They compared different pooling layers (Average Pool, Max Pool, SoftDBOW~\citep{softdbow-Philbin2008LostIQ}, NetFV~\citep{Lev2015RNNFV, Perronnin2015FisherVM, Sydorov2014DeepFK}, NetVLAD~\citep{netvlad-Arandjelovi2015} and NetRVLAD~\citep{netrvlad-Miech2017LearnablePW}) and video representation (I3D\citep{carreira2017quo}, C3D~\citep{c3d_2015}, and ResNet~\citep{He2015Resnet} features) with a sliding window approach at 0.5s stride (see Table~\ref{tab:methods-analysed-articles-classification}). The same pooling methods were investigated in~\citep{Vanderplaetse2020ImprovedSA}, where input was enriched with audio variables. The video was represented as ResNet features, and audio stream feature extractions were done with VGGish architecture (VGG~\citep{vgg-Simonyan2014VeryDC} pretrained on AudioSet~\citep{audioset-Gemmeke2017AudioSA}). It is worth mentioning that using modality fusion improved mAP of $7.43\%$ for the action classification task and $4.19\%$ for the action spotting task on the SoccerNet dataset. Experiments showed that mid-fusion was the most effective method ($73.7\%$), while early fusion achieved the worst performance ($64\%$). The result for late fusion is $68.4\%$. Similarly, the authors of \citep{NergrdRongved2021AutomatedED} conducted experiments with  multimodal models combining video and audio features in various settings. These experiments prove that combing modalities can lead to an improvement in model performance. They also acknowledged that the highest gain was observed in the classification of goal class, which can be associated with the audio reaction of supporters.
According to the authors of~\citep{Mahaseni2021SpottingFE}, enhancing event spotting may be achieved significantly by including short-range to long-range frame dependencies within an architecture. They have introduced a novel approach based on a two-stream convolutional neural network and Dilated Recurrent Neural Network (DilatedRNN) with Long Short-Term Memory (LSTM)~\citep{lstm_1997}. The Two-Stream CNN captures local spatiotemporal features required for precise details, while the DilatedRNN allows the classifier and spotting algorithms to access information from distant frames. \citet{soccer-self-attention-2020} used the self-attention mechanism to extract key frames and the NetVLAD network to obtain the temporal window-level (60s) features. The results of the classifier trained on the SoccerNet~\citep{soccernet} have improved from $67.2\%$ to $74.3\%$ accuracy by adding an attention mechanism.
ResNet3d pretrained on Kinetics-400~\citep{realtime-det-3d-rongved-2020} were found to be inferior to state-of-the-art models. However, the authors stated that this architecture is competitive in real-time settings or when the precise temporal localization of events is crucial.
A novel loss function CALF (Context-Aware Loss Function)~\citep{calf-Cioppa2020Context} for the segmentation model was introduced to improve action spotting model training. Instead of focusing on a single timestamp, CALF analyses the temporal context around the action. Frames are grouped into categories: far before, just before, just after, far after an action, and transition zones with associated parameters. Outputs of segmentation block feed spotting layer with YOLO-like loss. This architecture has significantly outperformed the baseline model ($+12.8\%$ of avg-mAP).

RMS Net~\citep{Tomei2021RMSNetRA} is a solution inspired by regression methods in object detection, combining classification and regression loss during the training. The model produces outputs that comprise both the probability of an action class and its corresponding temporal boundaries. The authors of this solution have also proposed masking procedures and implemented strategies to address the issue of data imbalance, which has led to an improvement in the mAP metric. They noticed that certain indicators of events tend to appear shortly after the event itself. By analyzing reactions, it is possible to infer whether an action has occurred or not. The masking procedure was constructed to focus only on frames occurring after an event which allows for a more targeted analysis of relevant video segments.  

\citet{9959985} implemented a Siamese neural network to conduct experiments of metric learning to detect soccer events. The most promising combination was Siamese Network with contrastive loss~\citep{koch2015siamese}, Efficient-NetB0~\citep{tan2021efficientnetv2}, and gated recurrent units (GRU)~\citep{gru-2014}.
In~\citep{Vats2020EventDI}, they introduced a multi-tower temporal convolutional neural network (MTTCNN) which considers that particular events occur with different frequencies in sports datasets. 

\citet{9106051} proposed new action classification dataset including SoccetNet~\citep{soccernet} videos. This article presents benchmarks for the classification of four actions (goal/shoot, yellow/red card, celebrate and pass) with I3D~\citep{carreira2017quo}, I3D-NL~\citep{wang2018non}, ECO~\citep{zolfaghari2018eco} and SlowFast~\citep{slowfast_Feichtenhofer2018SlowFastNF}.

After the release of SoccerNet-v2~\citep{soccernet-v2}, the action spotting task gained even more scientific interest~\citep{graph-for-action-spotting-cartas-2022, feature_combination_baidu_attention_xin_zhou_2021, ste-darwish-abdulrahman-2022, Multiple-Scene-Encoders-Shi-Yuzhi-2022, spotformer-cao-menggi-2022, multimodal-transformer-Gan-Yaozong-2022, Giancola2021TemporallyAwareFP, Hong2022SpottingTP, Soares2022TemporallyPA}. Published benchmarks reached the average mAP for tolerances ranging from 5s to 60s metric of $41.6$, and two years later, the result increased to $78.5$~\citep{AS_Dense_Detection_Anchors_RevisitedSoares2022}. Table~\ref{tab:methods-analysed-articles-spotting} presents results reported in analysed articles for action spotting task. It is worth noting that together with the increase in the number of classes (from $3$ in SoccerNet to $17$ in SoccerNet-v2), the task has been made more difficult. For instance, the performance of CALF decreased from $62.5$ on SoccerNet to $41.6$ on SoccerNet-v2.

\citet{Giancola2021TemporallyAwareFP} proposes a novel architecture known as NetVLAD++, which is based on NetVLAD pooling method. Similarly to \citep{calf-Cioppa2020Context}, they take into consideration both frames prior to the action (past) and frames subsequent to the action (future). The authors noted that certain actions share the same characteristics prior to the event but differ in what happens after the action. They provided the goal and shot as an example, highlighting that both actions have the same pre-event characteristics, but can be differentiated based on the events that follow the action. Future and past clips are processed independently as two levels of temporal context utilizing NetVLAD pooling layers. This approach surpasses the previous state of the art, notably outperforming traditional NetVLAD~\citep{soccernet-v2} by $22$ percentage points.

\citet{ste-darwish-abdulrahman-2022}~\footnote{\url{https://github.com/amdarwish/SoccerEventSpotting/tree/main/STE-v2}} presented two versions of Spatio-Temporal Encoder (STE) build by convolution layers, max pooling and fully connected layers. The architecture is distinguished by the speed of model training and low complexity  while maintaining good performance. The first version of the solution achieves $74.1\%$ avg-mAP on the test SoccerNet-v2 dataset and $40.1\%$ of tight avg-mAP. The modified model increased  tight avg-mAP to 58.5\%. Input frames are represented with Baidu features~\citep{feature_combination_baidu_attention_xin_zhou_2021}. Model's layers are divided into spatial and temporal representations. For the temporal encoder, they proposed three different time scales to capture events: [T, T/2, 1]. STE-v2 differs from STE-v1 in the last layer. The first version output predicted action class while STE-v2 returns a prediction of frame index in addition to the label.

After the development of transformer~\citep{NIPS2017_attention_is_all_you_need} neural networks in computer vision~\citep{Neimark2021VideoTN, Dosovitskiy2020AnII}, a large and growing body of literature has investigated
these architectures in action spotting task ~\citep{feature_combination_baidu_attention_xin_zhou_2021, Soares2022TemporallyPA, faster-tad-2022, multimodal-transformer-Gan-Yaozong-2022, spotformer-cao-menggi-2022, Multiple-Scene-Encoders-Shi-Yuzhi-2022, Transformer-Based-action-spotting-2022}.

Article~\citep{feature_combination_baidu_attention_xin_zhou_2021} suggested that video representation is crucial in automatic match analysis. Contrary to other solutions using ResNet and ImageNet features, they fine-tuned action recognition models such as TPN~\citep{Yang2020TemporalPN}, GTA~\citep{He2020GTAGT}, VTN~\citep{Neimark2021VideoTN}, irCSN~\citep{Tran2019VideoCW}, and I3D-Slow~\citep{slowfast_Feichtenhofer2018SlowFastNF} on SoccerNet-v2~\footnote{Features can be downloaded from \url{https://github.com/baidu-research/vidpress-sports}.} in semantic feature extraction block (this set of features will be referred as Baidu soccer features). Then, features from all extractors are concatenated into one vector. NetVLAD++ and transformer-based models are used to detect actions. Experiments showed that clip representation using all five feature types is much better than single-backbone. Transformer architecture with all features achieves $73.7\%$ of avg-mAP, while NetVLAD++ trained on the same representation achieves $74.1\%$ on test SoccerNet-v2. Furthermore, they proved that the proposed video representation is better than previously used ResNet features.

Inspired by work of~\citep{feature_combination_baidu_attention_xin_zhou_2021}, the study of \citep{spotformer-cao-menggi-2022} also represented video features as a fusion of multiple backbones. In addition to Baidu features, they chose action recognition models such as VideoMAE~\citep{Tong2022VideoMAEMA} and Video Swin Transformer~\citep{Liu2021VideoST} assuming that these models can extract task-specific features. On the top of each feature extractor is a multilayer perceptron that reduces dimensionality and retains main semantic information. Then, features are concatenated. The results of experiments show that applying dimensionality reduction improved the accuracy of the action spotter. The model consists of multiple stacked transformer encoder blocks that process each frame individually, followed by a feed-forward network to perform temporal segmentation. Finally, Soft-NMS~\citep{Bodla2017SoftNMSI} is used to filter predictions. The network of~\citep{feature_combination_baidu_attention_xin_zhou_2021} has been improved with the following adjustments: an additional input feature normalization, a learnable position embedding is used instead of sin-cos position encoding, changes of some hyperparameters values, and focal loss is applied to address data imbalance. Summing up the results, using only Baidu features achieved $56.9\%$ of tight avg-mAP, while enriching representation with VideoMAE and Video Swin Transformer increases the performance to $58.3\%$ of tight avg-mAP. The final model has an average-mAP equal to $76.1\%$ and a tight average-mAP equal to $60.9\%$ on the test set of SoccerNet-v2. 


\citet{Transformer-Based-action-spotting-2022} used transformer~\citep{Fan2021MultiscaleVT} to extract features from soccer videos. The architecture generated action proposals with a sliding window strategy. Then, the proposed clips are fed to a transformer-based action recognition module. Transformer representation is then processed by NetVLAD++\citep{Giancola2021TemporallyAwareFP}.

Multiple Scene Encoders architecture~\citep{Multiple-Scene-Encoders-Shi-Yuzhi-2022} uses a representation of multiple frames to spot action because some actions consist of subactions (e.g. goal can be represented as running, shooting and cheering). The paper reports $55.2\%$ Average-mAP using ResNet features, and $75.3\%$ with Baidu embedding features~\citep{feature_combination_baidu_attention_xin_zhou_2021}, once again showing a significant advantage thanks to the appropriate video representation.

Multimodal transformer-based model using both visual and audio information through a late fusion was introduced in~\citep{multimodal-transformer-Gan-Yaozong-2022} for action recognition. The transformer model is designed to capture the action's spatial information at a given moment and the temporal context between actions in the video. The input to the model consists of the raw video frames and audio spectrogram from the soccer videos. Video streams are processed with ViViT transformer~\citep{Arnab2021ViViTAV} and audio with a model based on Audio Spectrogram Transformer~\citep{Gong2021ASTAS}. Then, modalities representations are connected with the late fusion method as a weighted average of encoder results. Unlike the other articles modelling action spotting on the SoccerNet dataset, the authors of this article report results with the Top-1 Accuracy metric, so it is difficult to compare their results with other papers. However, the article has a broad analysis with reference to different architectures. The best analysed model trained exclusively on visual input achieved $60.4\%$, and the multimodal transformer proposed by~\citep{multimodal-transformer-Gan-Yaozong-2022} achieved  Top-1 Accuracy equal to $62.4\%$.

Action-spotting models commonly rely on using pretrained features as input due to the computation difficulties of end-to-end solutions. \citet{Hong2022SpottingTP} offers a more efficient, end-to-end architecture called E2E-Spot for accurate action spotting. Each video frame is represented by RegNet-Y~\citep{Radosavovic2020DesigningND} with Gate Shift Modules (GSM)~\citep{Sudhakaran2019GateShiftNF}. Then, similarly to~\citep{football-actions-tsunoda}, a recurrent network is used -- the resulting data sequence is modelled through a GRU network~\citep{gru-2014}, which creates a temporal context and generates frame-level class predictions.

As the development of action spotting solutions has progressed, the importance of accurate models capable of precisely localizing actions within untrimmed videos has become increasingly acknowledged~\citep{faster-tad-2022, Soares2022TemporallyPA, spotformer-cao-menggi-2022}. The performance of these models is commonly evaluated using the tight-avg-mAP metric, which measures their effectiveness within a specific small tolerance range (1, \dots, 5 seconds).

Inspired by Faster-RCNN~\citep{Ren2015FasterRCNN}, authors of \citep{faster-tad-2022} built architecture called Faster-TAD. Features are extracted using SwinTransformer~\citep{Liu2021SwinTH}. Then, similarly to the Faster-RCNN approach, 120 best proposals of action boundaries are generated. The boundary-based model was implemented to take into account the variance in action duration, as some actions may last only a few seconds while others may extend for several minutes. Then, the modules for correcting the proposals of action location and their classification work in parallel. The authors proposed an advanced context module consisting of three blocks (a Proximity-Category Proposal Block, a Self-Attention Block, and a Cross-Attention Block) to get semantic information for classification purposes. Proximity-Category Proposal Block gathers contextual information, a Self-Attention Block establishes relationships among proposals, and finally, a Cross-Attention Block gathers relevant context from raw videos concerning proposals. Architecture is enriched with Fake-Proposal Block for action boundary refinement and atomic features for better clip representation. They report $54.1\%$ of tight-avg-mAP, which is a $7.04\%$ gain compared to ~\citep{feature_combination_baidu_attention_xin_zhou_2021}.

Soares et al.~\citep{Soares2022TemporallyPA} also propose a solution that tackles the problem of imprecise temporal localization. The model returns detection confidence and temporal displacement for each anchor. The architecture consists of a feature extractor (ResNet-152 with PCA and Baidu soccer embeddings fine-tuned on SoccetNet-v2~\citep{feature_combination_baidu_attention_xin_zhou_2021}) followed by MLP. Then, features are processed by u-net~\citep{Ronneberger2015UNetCN} and transformer encoder. Additionally, they experimented with Sharpness-Aware Minimization (SAM)~\citep{sam-foret2021sharpnessaware} and mixup data augmentation~\citep{Zhang2017mixupBE}. Their solution significantly boosted performance with tight avg-mAP: from $54.1$ achieved by~\citep{faster-tad-2022} to $60.7$.
Then, they introduced improvements to this solution~\citep{AS_Dense_Detection_Anchors_RevisitedSoares2022} and won SoccerNet Challenge 2022. First, they modified preprocessing by resampling the Baidu embeddings~\citep{feature_combination_baidu_attention_xin_zhou_2021} to get greater frame frequency (2 FPS) and applying late fusion to combine them with the ResNet features. Also, soft non-maximum suppression (Soft-NMS)~\citep{Bodla2017SoftNMSI} was applied in the postprocessing step. These modifications resulted in a $4.4$ percentage point improvement over~\citep{Soares2022TemporallyPA} measured by tight avg-mAP.

Cioppa et al.~\citep{Cioppa2021CameraCA} conducted experiments exploring the utilization of camera calibration data in action spotting. The first phase involved the implementation of an algorithm based on  Camera Calibration for Broadcast Videos (CCBV) of~\citep{End-to-End-Camera-Calibration-2020-sha}. Results of Mask R-CNN~\citep{He2017MaskRcnn} model for object detection combined with camera calibration module allow preparing diverse feature sets, including top view representations with a 3D convolutional network, feature vectors representations (ResNet-34~\citep{He2015Resnet} and EfficientNet-B4~\citep{EfficientNet-2019}), and a player graph representation with graph convolutional network DeeperGCN~\citep{Li2020DeeperGCNAY}. In the graph, players are represented as nodes with edges connecting two players if the distance between them is less than $25$ meters. SoccerNet-v2 labels were divided into \textit{patterned} and \textit{fuzzy} groups based on prior knowledge of the predictive potential of player localization data for classifying these labels. Player localization plays a crucial role in the classification of \textit{patterned} classes (e.g. penalty, throw-in, cards) but is not relevant  for \textit{fuzzy} labels (substitution, ball out of play, foul). Two separate CALF~\citep{calf-Cioppa2020Context} networks were trained for each class group: one using calibration data to improve the classification of \textit{patterned} classes, and the other using only ResNet features for \textit{fuzzy} labels. They reported an avg-mAP of $46.8\%$, outperforming the unimodal CALF by $6.1$ percentage points.

Although the use of graphs had already found its application in sports analysis~\citep{Qi2020stagNetAA-volleyball, Passos2011NetworksAA-graph, Stckl2021MakingOP, Buld2018UsingN-graph}, Cioppa et al.~\citep{Cioppa2021CameraCA} were the first to use graph-based architecture to spot actions in untrimmed soccer videos. Then, 
Cartas et al.~\citep{graph-for-action-spotting-cartas-2022} developed another graph based-architecture that resulted in a substantial improvement over its predecessor, achieving an average mAP of $57.8\%$. Similarly to the previous solution, players are represented as nodes with attributes such as location, motion vector and label (player team 1/2, goalkeeper 1/2, referee), and are connected to another player by edge based on proximity (less than 5 meters). Players and referees are detected by the segmentation module PointRend~\citep{Kirillov2019PointRendIS}, and their position is projected onto a 2D pitch template through a homography matrix of camera calibration~\citep{Cioppa2021CameraCA}. After data cleaning, referees and players are classified using a convolutional network supported by a rule-based algorithm and clustering. Player information is enriched with motion vector represented by preprocessed optical flow extracted with FlowNet 2.0 CSS~\citep{Ilg2016FlowNet2E, flownet2-pytorch}. The model architecture consists of four dynamic edge graph CNN~\citep{Wang2018DynamicGC} blocks followed by NetVLAD~\citep{netvlad-Arandjelovi2015}  pooling layer. The authors experimented with multiple modalities and found that a graph-only model achieved $43.3\%$ of Average-mAP while adding video features increased the metric to $51.5\%$. Furthermore, incorporating audio features from VGGish
network~\citep{vgg-Simonyan2014VeryDC} with both video and graph streams resulted in an average mAP of $57.8\%$, surpassing both unimodal and bimodal methods.

\subsection{Spatio-Temporal Action Localization}

\begin{table*}
    \begin{center}
  \caption{Methods used for spatio-temporal action localization in analysed articles. \tikzcircle[green, fill=green]{3pt} means image.
  }
  \label{tab:methods-analysed-articles-spatio-temporal}
  \begin{tabular}{ccp{3.2cm}cccc}
    \toprule
    \textbf{Article} & \textbf{Dataset} & \textbf{Method} & \textbf{Features} & 
    \textbf{f@0.5} & \textbf{v@0.2} & \textbf{v@0.5} \\ 
    \midrule

     \citep{multisports-li-yixuan}  & MultiSports & ROAD~\citep{road_Singh2016OnlineRM} & \tikzcircle[green, fill=green]{3pt} & 
     3.90 & 0.00 & 0.00 \\
     \citep{multisports-li-yixuan}  & MultiSports & YOWO~\citep{yowo_Kpkl2019YouOW} & \tikzcircle[green, fill=green]{3pt} & 
     9.28 & 10.78 & 0.87 \\
  \citep{multisports-li-yixuan}  & MultiSports & MOC~\citep{moc_li_yixuan_2020} (K=7) & \tikzcircle[green, fill=green]{3pt} & 
  22.51 & 12.13 & 0.77 \\
   \citep{multisports-li-yixuan} & MultiSports &  MOC~\citep{moc_li_yixuan_2020} (K=11) & \tikzcircle[green, fill=green]{3pt} & 
   25.22 & 12.88 & 0.62 \\
      \citep{multisports-li-yixuan}  & MultiSports &  SlowOnly Det., 4 × 16~\citep{slowfast_Feichtenhofer2018SlowFastNF} (K=11) & \tikzcircle[green, fill=green]{3pt} & 
      16.70 & 15.71 & 5.50 \\
  \citep{multisports-li-yixuan}  & MultiSports &  SlowFast Det., 4 × 16~\citep{slowfast_Feichtenhofer2018SlowFastNF} (K=11) & \tikzcircle[green, fill=green]{3pt} & 
  27.72 & 24.18 & 9.65 \\

  \citep{Singh2022SpatioTemporalAD}  & MultiSports &  TAAD + TCN & \tikzcircle[green, fill=green]{3pt} & 
  \textbf{55.3} & - & \textbf{37.0}  \\
  \citep{Faure2022HolisticIT}  & MultiSports & HIT  & \tikzcircle[green, fill=green]{3pt} &  
  33.3 & \textbf{27.8} & 8.8 \\

  \bottomrule
\end{tabular}
\end{center}
\end{table*}

The release of the MultiSports dataset~\citep{multisports-li-yixuan} for spatio-temporal localization of multiple sportsmen may contribute towards a better understanding of actions performed by individual players.  Approaches to addressing this challenge can be categorized into  frame-level and clip-level models~\citep{multisports-li-yixuan}. The frame-level models predict the bounding box and action type for each frame and then integrate these predictions. Conversely, clip-level methods, also called action tubelet detectors, endeavour to model both temporal context and action localization. Authors of the MultiSports dataset published benchmarks for the proposed task training 
frame-level models (ROAD~\citep{road_Singh2016OnlineRM}, YOWO~\citep{yowo_Kpkl2019YouOW}) and clip-level models (MOC~\citep{moc_li_yixuan_2020}, SlowOnly~\citep{slowfast_Feichtenhofer2018SlowFastNF} and SlowFast~\citep{slowfast_Feichtenhofer2018SlowFastNF}). Results of experiments are summarized in Table~\ref{tab:methods-analysed-articles-spatio-temporal}.

ROAD~\citep{road_Singh2016OnlineRM} is an algorithm for real-time action localization and classification which uses the Single Shot Multibox Detector (SSD)~\citep{ssd_Liu2015SSDSS} method to independently detect and classify action boxes in each frame, without taking into account temporal information. Afterwards, the predictions from each frame are combined into action tubes through a novel algorithm. Similarly, You Only Watch Once (YOWO)~\citep{yowo_Kpkl2019YouOW} method for identifying actions in real-time video streams links results from individual frames into action tubes through a dynamic programming algorithm. It uses two concurrent networks: a 2D-CNN to extract spatial features from key frames and a 3D-CNN to extract spatio-temporal features from key frames and preceding frames. Then, the features from these two networks are combined through a channel fusion and attention mechanism and fed into a convolution layer to predict bounding boxes and action probabilities directly from video clips. Another approach was proposed in article~\citep{moc_li_yixuan_2020} introducing Moving Center Detector (MOC detector). It models an action instance as a series of moving points and leverages the movement information to simplify and enhance detection. The framework consists of three branches: (1) Center Branch for detecting the center of the action instance and action classification, (2) Movement Branch for estimating the movement between adjacent frames to form a trajectory of moving points, and (3) Box Branch for predicting the size of the bounding box at each estimated center. They return tubelets, which are then linked into video-level tubes through a matching process. SlowFast~\citep{slowfast_Feichtenhofer2018SlowFastNF} comprises of two parallel branches. The slow branch identifies spatial semantics that exhibits minimal fluctuations, thus allowing for a low frame rate. Conversely, the fast branch is responsible for detecting rapid changes in motion, requiring a high frame rate to operate effectively. During training, data from the fast branch is fed to a slow neural network and at the end, the results of the two networks are concatenated into one vector. Faster R-CNN~\citep{Ren2015FasterRCNN} with a ResNeXt-101-FPN~\citep{Lin2016FeaturePN, Xie2016AggregatedRT} backbone was used to detect people. As the name suggests, the SlowOnly model uses only the slow path of SlowFast.
Table~\ref{tab:methods-analysed-articles-spatio-temporal} summarizes the results and indicates that the SlowFast detector achieved the best performance within benchmark models. Metrics are computed for all sports, not only for soccer.

The results obtained by Gueter Josmy Faure1 et al. in~\citep{Faure2022HolisticIT}~\footnote{\url{https://github.com/joslefaure/HIT}} suggest that including pose information can be valuable to predict actions. Authors motivate their architecture with the fact, that actions can be defined as interactions between people and objects. Their multimodal Holistic Interaction Transformer Network (HIT) fusing a video stream and a pose stream surpasses other models on the MultiSports dataset. Each stream composes of person interaction, object interaction and hand interaction to  extract action patterns. For each modality, Intra-Modality Aggregator (IMA) facilitates learning valuable action representations. Then, an Attentive Fusion Mechanism (AFM) is utilized to merge the various modalities, retaining the most significant features from each modality. 3D CNN backbone~\citep{slowfast_Feichtenhofer2018SlowFastNF} processes video frames, Faster RCNN~\citep{Ren2015FasterRCNN} with ResNet-50-FPN~\citep{Xie2016AggregatedRT, Lin2016FeaturePN} backbone predict bounding boxes and spatio transformer~\citep{Zheng20213DHP} is pose encoder. This method outperformed others in terms of f-mAP@0.5 and v-mAP@0.2.

In existing solutions to spatio-temporal action recognition, tube detection involves extending a bounding box proposal at a keyframe into a 3D temporal cuboid and pooling features from nearby frames. However, this approach is not effective when there is significant motion. The study~\citep{Singh2022SpatioTemporalAD} propose cuboid-aware feature aggregation to model spatio-temporal action recognition. Also, they improve actor feature representation through actor tracking data and temporal feature aggregation along the tracks. The experiments show that the proposed method called Track Aware Action Detector (TAAD) outperforms others, especially for large-motion actions.

\subsection{Summarizing Multimodal Action Scene Understanding}

Multimodal machine learning is a powerful approach combining different pieces of information to understand complex phenomena comprehensively. Fusing information from multiple modalities leads to a deeper comprehension of the underlying processes, enabling superior predictive performance compared to unimodal models. In the realm of soccer, the adoption of multimodal approaches has successfully improved the accuracy of predictive models through the integration of diverse sources of data and the extraction of more meaningful insights~\citep{Vanderplaetse2020ImprovedSA, NergrdRongved2021AutomatedED, graph-for-action-spotting-cartas-2022, Cioppa2021CameraCA, feature_combination_baidu_attention_xin_zhou_2021}. 

Classical methods of action recognition relied mostly on data preparation and feature engineering. Thus, authors extracted different features from video clips, including logo frame detection, audio MPEG descriptors, camera motion, zoom indicator, colour layout, dominant colour, referee’s whistle indicator, view category, and Histogram of Oriented Gradients (HoG)~\citep{1467360}. It is worth noting that these methods widely used a combination of audio and visual features~~\citep{fuzzy_HOSSEINI2013846, bayesian_network_2015, Raventos2015-vt, kapela2015real, 1199134, 1221333, 1221608}. The more recent methods have primarily relied on visual embedding alone~\citep{soccernet-v2, feature_combination_baidu_attention_xin_zhou_2021, calf-Cioppa2020Context, Tomei2021RMSNetRA}. However, using a fusion of multiple representations of a single source (e.g. Baidu embeddings for video~\citep{feature_combination_baidu_attention_xin_zhou_2021}), that can also be considered as multimodality, has proven to be more effective than using a single model to represent video (e.g. ResNet features). The release of Baidu soccer embeddings has resulted in researchers favouring them over ResNet features originally presented  by SoccerNet authors. Moreover, Baidu embeddings were further extended by two additional models~\citep{spotformer-cao-menggi-2022}. Experiments showed that incorporating audio streams~\citep{Vanderplaetse2020ImprovedSA, NergrdRongved2021AutomatedED, graph-for-action-spotting-cartas-2022}, graph networks~\citep{graph-for-action-spotting-cartas-2022, Cioppa2021CameraCA} and optical flow~\citep{graph-for-action-spotting-cartas-2022} can also provide significant value to the model. In the case of spatio-temporal action localization models, the fusion of a video stream and a pose stream surpassed other solutions~\citep{Faure2022HolisticIT}.









\section{Discussion}
\label{sec:discussion}

Automatic action recognition and localization is relevant from the perspectives of many in the soccer industry: coaches, scouts, fans and broadcasters. The Internet provides various sources of information on match results, highlights, and non-structured data. Also, many matches are recorded via different cameras, and TV and radio provide audio commentary to some matches. Therefore, it seems that soccer can be an excellent source of multimodal data. However, collecting high-quality and realistic soccer videos is complex and challenging for several reasons.

\paragraph{Datasets} One of the main challenges in gathering annotated soccer data is that it can be challenging to obtain the necessary data due to licences and limited public availability of broadcast content. An even bigger challenge is assessing data for smaller or less popular leagues. Preparing annotated data is a laborious, time-consuming and expensive task because it involves manual annotation of video footage. This process may require a team of trained analysts to watch and annotate every match. The quality of the annotations may depend on the level of expertise of the analysts, which can further affect the accuracy of the data. To avoid this, a match is sometimes annotated by several people, with the recorded data being cross-referenced. Nevertheless, the interpretation of any event is always subjective and open to numerous interpretations. For instance, two analysts may disagree on whether a particular incident should be classified as a foul or not. Also, obtaining accurate frame level annotations of events is difficult due to inaccuracies in defining the beginning and end of the action. This subjectivity can result in inconsistencies in the data and make it difficult to compare or analyze different datasets.

To be useful and informative, soccer data must meet certain requirements that ensure its quality, reliability and usefulness to the industry. Annotations should be prepared in a standardized manner to ensure comparability and consistency. Firstly, current sport datasets provide access to trimmed short clips~\citep{ucf-sports-0-dataset-soomro-khurram}, but this assumption is unrealistic. To apply models to real scenarios, they should be trained on whole untrimmed videos, such as SoccerNet~\citep{soccernet, soccernet-v2}. Secondly, broadcast-like videos with moving camera and replays are the most natural to collect. It could be difficult and costly to gather videos from multiple cameras for the same scene; however, middle camera recordings or drone recordings could be valuable and useful for everyday training, even being affordable for smaller clubs. 

Although there are already quite a few soccer datasets, there is still room for improvement. Action recognition is essential in the analysis of matches in lower leagues to find talented players. Unfortunately, many games are not broadcasted, and the installation of many cameras on the pitch is too expensive. Thus, datasets consisting of drone recordings or middle camera-only recordings and models trained on them could supplement the work of scouts. 
Moreover, even though there are a lot of multimodal soccer data sources, there is a lack of publicly available datasets, including them. 
The development of spatio-temporal action localization methods in soccer can lead to the easy highlighting of actions performed  by individual players. Furthermore, combining results with homography allows statistics per action to be computed. For instance, \textit{"Player [] covers a distance [] while dribbling with an average speed of []"}. Despite the wide range of applications, there is no dedicated soccer dataset for this task. In contrast, authors of~\citep{volleyball-spatio-temporal-2016} proposed a widely used volleyball dataset consisting of sportsman position and action annotation along with group activity labels.

\paragraph{Methods and Potential of Multimodality}

Soccer action scene understanding, which can be divided into action classification, spotting and spatio-temporal action localization, is a crucial aspect of analyzing soccer matches. The mentioned tasks vary in difficulty, with action classification being the easiest and action spotting being a more complex task that involves classification and finding temporal localization of actions. Spatio-temporal action localization makes the task more difficult by adding the spatial aspect.
Considering the temporal context of actions is essential for several reasons.  Firstly, some actions follow each other and information that one occurred should increase the probability of the associated action. For instance, yellow cards occur after fouls, which are then followed by free kicks. Moreover, analysis of frames before and after an action can provide valuable information about the occurrence of actions. Before a goal-scoring situation, one team runs towards the opposite goal in the video, and the audio track can capture the reactions of reporters and fans, including cheers or boos. Also, the results of an action can be deduced from what happens after the event. If there is a goal, players and their fans celebrate it. 
It can be challenging to train models to localize actions when sudden position changes and fast movement occur during attacks and dribbling. The duration of events can also vary significantly, such as the difference between the time taken for a yellow card and the ball being out of play.

Model performance on benchmark datasets has fluctuated over the past few years. Action spotting on SoccerNet~\citep{soccernet} has gained about 17 percentage points of average mAP in comparison to the baseline (from 49.7 to 66.8). Similarly, after the release of SoccerNet-v2~\citep{soccernet-v2} in 2021, containing 17 action classes, the average mAP increased from 18.6 to 78.5. A similar trend is observed on the MultiSports~\citep{multisports-li-yixuan} dataset for spatio-temporal action localization, where the best model published by dataset's authors was outperformed in terms of frame-mAP@0.5 (almost twice as good) and in terms of video-mAP@0.5 (four times as good)~\citep{Singh2022SpatioTemporalAD}. 
Huge improvements were made thanks to enriching data representation with other data sources, such as audio~\citep{Vanderplaetse2020ImprovedSA, NergrdRongved2021AutomatedED, graph-for-action-spotting-cartas-2022}, graphs~\citep{graph-for-action-spotting-cartas-2022, Cioppa2021CameraCA}, and pose~\citep{road_Singh2016OnlineRM}. Although SoccerNet includes reporter's commentary track, text input has not yet been used in modelling. A textual input can provide a plethora of valuable information as sports commentators describe the unfolding events on the pitch. Thus, work on the enriching of representation with text data is promising; experiments will be needed to verify this conjecture. Multimodal models can also be defined as a different representation of the same input, e.g. concatenation of various text embeddings. This approach was used in many articles~\citep{feature_combination_baidu_attention_xin_zhou_2021, Multiple-Scene-Encoders-Shi-Yuzhi-2022, spotformer-cao-menggi-2022, Soares2022TemporallyPA, AS_Dense_Detection_Anchors_RevisitedSoares2022}. The most groundbreaking is the article of Baidu Research~\citep{feature_combination_baidu_attention_xin_zhou_2021} which published multiple semantic features of action recognition models, which were then used in other articles.

To sum up, action scene understanding in soccer has attracted much attention from research teams in recent years. Many publicly available datasets have been released, and models have improved the accuracy of action spotting and recognition. Nevertheless, some interesting and relevant problems remain to be addressed, including spatio-temporal action localization datasets and models dedicated to soccer and experiments with multimodal data such as textual commentary.


\section*{Acknowledgments}
We would like to thank Olaf Skrabacz and Filip Boratyn for his valuable insights, comments and recommendations. This research was supported by NASK - National Research Institute.

\section*{Declarations}

\subsection*{Availability of data and materials}
Not applicable

\subsection*{Code availability}
Not applicable

\bibliography{references}

\end{document}